\documentclass[lettersize,journal]{IEEEtran}
\usepackage{amsmath,amsfonts}
\usepackage{algorithmic}
\usepackage{algorithm}
\usepackage{array}
\usepackage[caption=false,font=normalsize,labelfont=sf,textfont=sf]{subfig}
\usepackage{textcomp}
\usepackage{stfloats}
\usepackage{url}
\usepackage{verbatim}
\usepackage{graphicx}
\usepackage{cite}
\newsavebox{\picbox}
\newcommand{\imageWithCorners}[3]{
\savebox{\picbox}{\includegraphics[width=#2]{#3}}%
\tikz\node[
  draw,rounded corners=#1,
  line width=0pt, color=white,
  minimum width=\wd\picbox,
  minimum height=\ht\picbox,
  path picture={\node at (path picture bounding box.center) {\usebox{\picbox}}; }]
{};}
\usepackage[capitalize,noabbrev]{cleveref}
\usepackage{subfig}  %
\usepackage[textsize=tiny]{todonotes}
\usepackage{booktabs} %

\begin{document}

\title{Learning to Generalize with Object-centric Agents\\ in the Open World Survival Game Crafter}

\author{
Aleksandar Stani\'{c}\textsuperscript{† \ 1}, Yujin Tang\textsuperscript{2}, David Ha\textsuperscript{2}, J\"{u}rgen Schmidhuber\textsuperscript{1 \ 3}
\thanks{† The majority of this work was carried out as a student researcher at Google Brain. \textsuperscript{1} IDSIA, USI, SUPSI, Lugano, Switzerland, \textsuperscript{2} Google Brain, \textsuperscript{3} AI Initiative, KAUST, Thuwal, Saudi Arabia.}%
\thanks{A shorter version of this paper appeared at the Decision Awareness in Reinforcement Learning and Responsible Decision Making in Dynamic Environments Workshops at ICML 2022.}
}

\markboth{}%
{Shell \MakeLowercase{\textit{et al.}}: A Sample Article Using IEEEtran.cls for IEEE Journals}

\maketitle

\begin{abstract}
Reinforcement learning agents must generalize beyond their training experience.
Prior work has focused mostly on identical training and evaluation environments.
Starting from the recently introduced Crafter benchmark, a 2D open world survival game, we introduce a new set of environments suitable for evaluating some agent's ability to generalize on previously unseen (numbers of) objects and to adapt quickly (meta-learning).
In Crafter, the agents are evaluated by the number of unlocked achievements (such as collecting resources) when trained for 1M steps.
We show that current agents struggle to generalize, and introduce novel object-centric agents that improve over strong baselines.
We also provide critical insights of general interest for future work on Crafter through several experiments.
We show that careful hyper-parameter tuning improves the PPO baseline agent by a large margin and that even feedforward agents can unlock almost all achievements by relying on the inventory display. 
We achieve new state-of-the-art performance on the original Crafter environment.
Additionally, when trained beyond 1M steps, our tuned agents can unlock almost all achievements.
We show that the recurrent PPO agents improve over feedforward ones, even with the inventory information removed. 
We introduce CrafterOOD, a set of 15 new environments that evaluate OOD generalization.
On CrafterOOD, we show that the current agents fail to generalize, whereas our novel object-centric agents achieve state-of-the-art OOD generalization while also being interpretable. 
Our code is public.\footnote{https://github.com/astanic/crafter-ood}
\end{abstract}


\section{Introduction}

\IEEEPARstart{C}{ommon} benchmarks and solid baselines are essential for developing new models and correctly measuring progress in machine learning.
Datasets for supervised learning (such as ImageNet~\cite{deng2009imagenet}) and game-based environments for reinforcement learning (RL) (such as Atari~\cite{bellemare13arcade}, ProcGen~\cite{cobbe2020leveraging} and MineRL~\cite{guss2019neurips,guss2021minerl}) have played a crucial role in developing new methods and improving the state of the art.
To judge the improvements offered by novel methods, it is critical to compare them to solid yet simple baselines,
especially in deep RL, where reproducing existing work is often hard due to a plethora of difficulties \cite{henderson2018deep}.

Early benchmarks were important to show that certain deep RL methods can learn control from high-dimensional images, e.g., neuroevolution via compressed network encodings \cite{koutnik2013evolving} on the TORCS benchmark \cite{wymann2000torcs}, or the DQN agent \cite{mnih2015human} on the Atari benchmark \cite{bellemare13arcade}.
These benchmarks, however, focused on narrow task sets, and conducted training and evaluation on in-distribution environments, where the agents could (simply) memorize sequences of actions leading to high reward without \textit{understanding} the underlying mechanics of the world. 
Intelligent agents should discover salient objects (entities) of the world, identify relevant properties of objects (e.g., shape is important, color is not) and ignore irrelevant parts of the environment.
For better evaluation, the focus has recently shifted to studying agent performance on carefully designed benchmarks~\cite{osband2020bsuite}, with emphasis on agents that generalize to environments beyond what they are trained on (e.g., ProcGen or Crafter~\cite{hafner2021benchmarking}).

\begin{figure}[t]
	\centering
	\includegraphics[width=\linewidth]{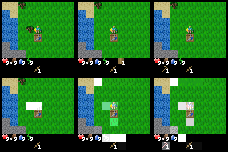}
	\caption{\it 
	Crafter gameplay showing an agent collecting resources (wood) and crafting a weapon (wooden pickaxe), and its attention.
	Top row: input images. Bottom row: visualized attention. Episode steps are plotted horizontally.}
	\label{fig:front-page}
\end{figure}

Here we focus on Crafter~\cite{hafner2021benchmarking}, a recently introduced open world survival game that allows tractable investigation of new agents and their generalization, exploration, and long-term reasoning capabilities.
Compared to other benchmarks, Crafter has a set of advantages that make it suitable for RL research: fast iteration speed (agents can be trained within a few hours on a standard GPU and a single CPU core), model evaluation by inspecting semantically meaningful achievements unlocked by the agents, and controllable environmental objects that facilitate systematic studies.
Our experiments on the original Crafter environment offer important, previously unpublished insights into baselines and environmental workings.
In addition, we introduce two sets of new environments, CrafterOODapp and CrafterOODnum, that test out-of-distribution (OOD) generalization to unseen types and numbers of objects, respectively.
This also sets the stage for developing fast-adaptation algorithms in the context of meta-learning.

\begin{figure*}[ht]
	\hspace{7mm}
	\subfloat[]
	{
		\centering
		\includegraphics[width=.23\linewidth]{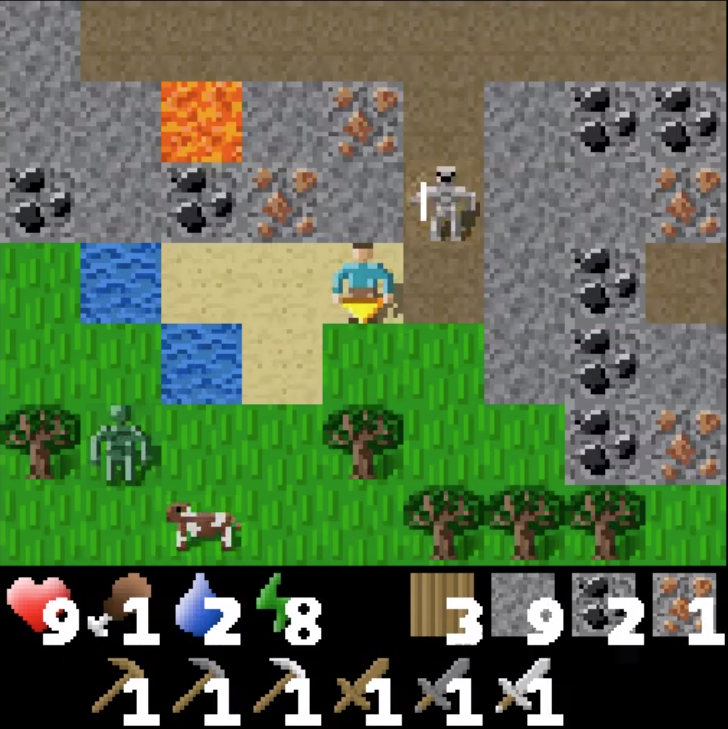}
		\label{sfig:models1}
	}
	\hspace{7mm}
	\subfloat[]
	{
		\centering
		\includegraphics[width=.22\linewidth]{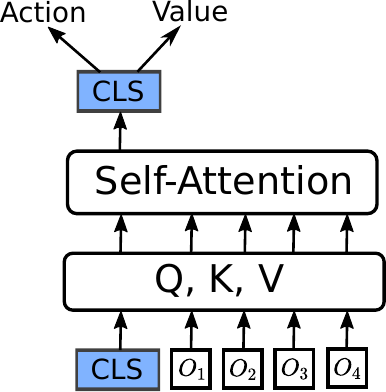}
		\label{sfig:models2}
	}
	\hspace{7mm}
	\subfloat[]
	{
		\centering
		\includegraphics[width=.38\linewidth]{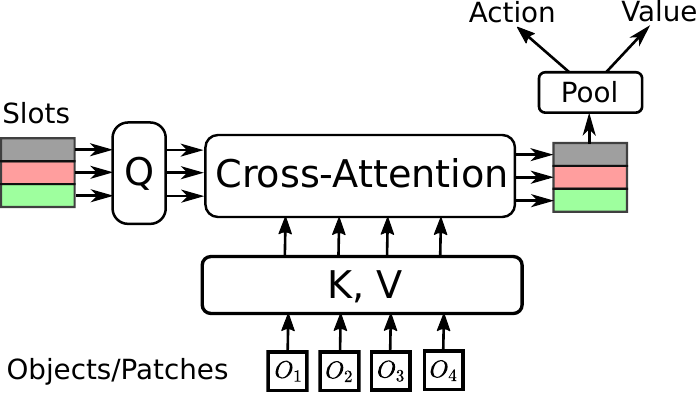}
		\label{sfig:models3}
	}
	\caption{\it 
	(a) Crafter gameplay. (b) Object-centric Self-Attention (OC-SA) model with a CLS token and queries (Q), keys (K) and values (V). (c) Slot-based Object-centric Cross-Attention (OC-CA) model. Pool operation averages the slots to produce the final input vector fed into the actor and critic networks.
	}
	\label{fig:models}
	\vspace{-3mm}
\end{figure*}

Our main contributions are:
(1) We show that careful hyper-parameter tuning improves the PPO baseline agent by a large margin. We achieve new state-of-the-art performance on the original Crafter environment.
(2) We show that feedforward agents can unlock almost all achievements by relying on the inventory display (bottom part of the images in \cref{fig:front-page,sfig:models1}) -- a type of ``scratchpad'').
(3) We show that recurrent agents improve over feedforward ones, even when the inventory information is  removed.
(4) We show that tuned agents can unlock almost all achievements.
(5) We introduce CrafterOOD, a set of 15 new environments, to evaluate OOD generalization of agents.
(6) We introduce novel object-centric agents that achieve state-of-the-art OOD generalization while being interpretable.

\section{Environments}

Here we briefly summarize essential properties of the Crafter environment and the newly introduced CrafterOODapp and CrafterOODnum environments. 
For additional details on Crafter we refer to \cite{hafner2021benchmarking}.

\textbf{Crafter.}
Crafter is an open world survival game for RL research whose game dynamics are inspired by the popular game Minecraft.
The benchmark is designed to facilitate existing research challenges: procedural generation to test generalization, exploration due to a deep technology tree of achievements conditioned on one another, high-dimensional image observations requiring representation learning, memory through partial observability, and finally, sparse rewards that require long-term reasoning and credit assignment.
It has the right level of complexity that is challenging enough for the current agents to unlock all achievements, yet not overly complex such that it facilitates interpretability of the agent's behavior and evaluation by unlocking semantically meaningful achievements.
In Crafter, a unique terrain is generated for every episode with grasslands, lakes, and mountains that contain forests, caves, ores, and lava.
The world consists of $64 \times 64$ grid cells, but the agent only observes $7 \times 9$ grid cells making Crafter a partially observed environment.
At each step the agent receives a $64 \times 64 \times 3$ input image as in \cref{sfig:models1}, and returns an action from a categorical space of 17 actions.
Of those actions, four are for moving in the environment, one is for sleeping, and the rest are for crafting or interacting with specific tools or resources.
For an example of gameplay, see Figure~\ref{sfig:models1}, where you can see the player (agent) in the middle, a skeleton on its top right, a zombie bottom-left from the agent, a cow, water, trees, stones, iron and lava.

In the bottom-left part of \cref{sfig:models1} the agent's health status is shown.
The heart represents the number of lives it has left, and next to it, the levels of food, water, and energy (that it recharges by sleeping).
Next to this is the inventory display that shows the resources (here wood, stone, coal, and iron) and tools and weapons (here wooden, stone, and iron pickaxes and swords) it has collected or crafted so far.
The agent must collect resources such as wood, stone, coal, and iron to craft tools.
The agent needs to eat food, drink water, sleep and defend against enemies to prevent its health from reaching zero.
All these agents and resources are represented as objects in the environment.
Eating food, collecting resources, defeating enemies, or crafting new tools are all ``achievements''.
The first time the agent unlocks an achievement in an episode it receives a reward of $1$.
On the other hand, for every lost health point the reward is $-0.1$, and for every regained health point $+0.1$.
When the health reaches zero or when stepping into lava the agent dies, and the episode is over.
Every time the agent unlocks an achievement previously unlocked in the episode, it does not receive any further reward.
This should encourage the agent to explore and collect  as many achievements as possible instead of exploiting the ones it already knows.
Collecting the diamond is the final and the most difficult achievement.
To collect a diamond, the agent must build several tools that require even more basic tools and resources to be collected.
This results in a very deep technology tree (see Figure 4 in the Crafter paper~\cite{hafner2021benchmarking}).

In total, there are $22$ achievements and the main evaluation metric is the so-called \textit{crafter score} $S$. 
It is computed by averaging the unlocked achievements in the log space (to account for differences in their difficulties): 
$S=\exp(\frac{1}{N}\sum_{i=1}^N\ln(1+s_i))$, where $s_i\in[0,100]$ is percentage of episodes in which the achievement $i$ was unlocked and $N$ is the number of achievements (here $N=22$).
This score is invariant to the number of times an achievement was unlocked in an episode (once it is achieved).
Due to averaging in the log space, the score will increase more for unlocking difficult achievements in a few episodes (e.g. a diamond) than for unlocking easy ones more frequently (e.g. collecting wood).
The final evaluation metrics is the geometric mean of $S$ over all episodes.

\textbf{CrafterOODapp.}
Building on top of Crafter, we introduce CrafterOODapp, where we develop another axis of variation: object appearance based on color.
We introduce new variants of objects the agent most frequently interacts with: trees, cows, zombies, stones, coal, and skeletons.
For a detailed overview of the newly introduced objects, see Figure~\ref{fig:textures-ood}.
This evaluates generalization to unseen or infrequently seen objects.
The agent is trained on an environment with one distribution of such objects' appearances and then evaluated on an environment with a different distribution, possibly one with objects never seen during training. 
In this way, CrafterOODapp sets the stage for the development of agents based on fast-adaptation or meta-learning.

In the default Crafter environment, the agent always sees only the first variant of an object.
On the other hand, in CrafterOODapp, the agent encounters one of the four object variants with varying frequencies of occurrence.
For example, in the easiest scenario, the agent sees all four object variants ($O_1,O_2,O_3,O_4$) with equal ($25\%$) probability in the training environment, and then in the evaluation environment it encounters only the ``last'' three objects $O_2,O_3,O_4$ with equal (but increased, $33.3\%$) probability.
We then generate progressively more challenging scenarios where the training environment contains more instances of object variant $O_1$.
Lastly, in the most extreme case of zero-shot generalization, we train agents on environments containing only the first object $O_1$ and then evaluate them on environments containing only the last three objects $O_2,O_3,O_4$.

\textbf{CrafterOODnum.}
Also, building atop Crafter, we introduce CrafterOODnum, where the agent encounters different numbers of objects during the training and evaluation phases.
As in CrafterOODapp, we vary the objects the agent most frequently interacts with: trees, coal, cows, zombies, and skeletons.
To get a clear picture of these environments, in \cref{tbl:crafterood-numbers} we show the numbers of each of these objects in the respective environment variants.
Starting from default object distributions in Crafter, we increase or decrease object numbers by powers of two.
These variants are particularly challenging as the agent might face never-before-seen situations such as fighting against more enemies or surviving in environments with resources scarcer compared to what it was trained on.
Importantly, changing the number of objects does not change the maximum score the agent can obtain, as it is still possible to unlock all the achievements, but with different difficulties.
However, this changes the reward function and makes the credit assignment problem more difficult because reward is sparser with fewer objects.

\section{Methods}

In this section, we describe the agents' network architectures. 
We first introduce CNN feedforward baselines, followed by LSTM-based recurrent agents, and finally introduce object-centric agents.
To optimize agents we use the PPO \cite{schulman2017proximal} implementation in the stable baselines \cite{stable-baselines3}.

\subsection{Linear and Recurrent PPO}

PPO learns to map images to actions via policy gradients.
Two feedforward variants are investigated that differ by the CNN policy they use.
We introduce a recurrent version based on an LSTM ~\cite{hochreiter1997long} to account for partial observability.
Note that the first application of policy gradients to LSTM  dates back to 2007 \cite{wierstra2007solving}.
For architecture details see Appendix~\ref{sec:appendix:hp-heatmaps}.

\textbf{PPO with NatureCNN (PPO-CNN).} 
This baseline is architecturally identical to the one used in \cite{hafner2021benchmarking}, with the CNN policy from the DQN paper \cite{mnih2015human}.

\textbf{PPO with size-preserving CNN (PPO-SPCNN).} 
Size-preserving CNN (SPCNN) differs from the PPO-CNN baseline by the CNN architecture.
We introduce it as a baseline for object-centric agents that use SPCNN for ``feature mixing,'' inspired by SlotAttention's visual encoder~\cite{locatello2020object}.
SPCNN does not have pooling layers, so the resulting output tensor is of the same height and width as the input image.
The policy input tensor is much larger (64x64x64 instead of 8x8x64 for CNN) making the policy also very large (134M vs. 1M parameters).

\textbf{Recurrent PPO Agents.}
Crafter is a partially observable environment.
The agent observes only a part of the world (see Figure~\ref{sfig:models1}).
To perform well, it needs to remember the locations of resources, e.g. food, mining materials, and where it placed objects for crafting new tools, e.g. table or furnace.
Additionally, some achievements require the agent to perform a long chain of reasoning.
For example, to collect a diamond, the agent needs to have an iron pickaxe, for which it needs to have crafted a furnace and collected coal, for which in turn it needs to have previously collected wood, placed a table, made a wooden pickaxe, and collected stone (in that particular order).
For the complete overview of the technology tree, see Figure 4 in \cite{hafner2021benchmarking}.
For these reasons, we introduce recurrent agents (LSTM-CNN and LSTM-SPCNN) where we use LSTMs as the critic and the actor networks.
These networks, in theory, entitle the agent to have memories of the world map and its inventory and can help unlock achievements.

\subsection{Object-centric Agents}
Our experiments show that the baseline agents fail to generalize to OOD environments.
This is in line with the previous findings that neural networks fall short in generalization.
One hypothesis for this is due to their inability to dynamically and flexibly bind information distributed throughout the network \cite{greff2020binding}, also known as the binding problem  \cite{von1994correlation, roskies1999binding}.
To address this issue, there has been an increased interest in designing object-centric neural networks that learn (discrete) object representations from raw visual input, which support efficient learning and generalization to novel scenarios and behaviors.
Object-centric methods have successfully been used for OOD generalization in supervised and unsupervised learning \cite{greff2017neural,van2018relational,kosiorek2018sequential,stanic2019rsqair,greff2019multi,locatello2020object,stanic2020hierarchical,kipf2021conditional} and in RL \cite{watters2019cobra,veerapaneni2020entity,kipf2020Contrastive,carvalho2021feature}.
Since Crafter is composed of objects, we expect these methods to facilitate learning and show stronger generalization capabilities. 
In this vein, we designed two object-centric agents and investigated them on CrafterOOD environments.
They take as input either patches extracted from the image through a size-preserving CNN (Figure~\ref{fig:attn2}) or learn their own representation of an object by attending over the whole input tensor (Figure~\ref{fig:attn-learned}).

\textbf{Object-centric Self-Attention agents (OC-SA)} (Figure~\ref{sfig:models2} and \cref{tbl:OC-SA}) learn a policy via a dot-product self-attention between the input patches and a learned ``CLS'' token (a vector initialized to unit Gaussian parameters and optimized via backprop).
This is similar to using the CLS token in BERT \cite{devlin2018bert}.
Let $(x_1, ..., x_k, CLS)\in\mathbb{R}^{(k+1) \times d_{in} }$ be the input sequence, where $k$ is the number of input patches and $d_{in}$ is their dimensionality.
Dot-product self-attention is defined as:
\begin{equation}
Attention(Q, K, V) = softmax \Big(\frac{QK^T}{\sqrt{d}}\Big)V, 
\label{eq:attn}
\end{equation}
where $Q\in\mathbb{R}^{(k+1) \times d}$, $K\in\mathbb{R}^{(k+1) \times d}$, $V\in\mathbb{R}^{(k+1) \times d}$, are the query, key and value matrices, resulting from a linear mapping of the input sequence onto a space of dimension $d$.
Absolute sinusoidal positional embeddings \cite{vaswani2017attention} are added to the input, enabling the learning of relative patch positions, e.g. if an enemy is nearby.

\textbf{Object-centric Cross-Attention agents (OC-CA)} (Figure~\ref{sfig:models3} and \cref{tbl:OC-CA}) learn a set of vectors, which we refer to as 'slots', by attending either to a patch grid or the whole input (this can be seen as an extreme case of the patch grid with patch size of one pixel).
This idea was introduced in SetTransformer \cite{lee2019set} and successfully used for object detection \cite{carion2020end}, unsupervised learning of objects \cite{locatello2020object}, learning permutation-invariant agents \cite{tang2020neuroevolution,tang2021sensory} and general perception modules \cite{jaegle2021perceiver,jaegle2021perceiverio,alayrac2022flamingo}.
OC-CA computes attention between queries, keys and values as in self-attention \cref{eq:attn}, but with the queries coming from the $n$ learned slots: $Q\in\mathbb{R}^{n \times d}$, $K\in\mathbb{R}^{k \times d}$, $V\in\mathbb{R}^{k \times d}$, where $k$ is the number of input patches or image pixels for \textit{learned} objects.
The extreme patch size allows for higher expressiveness as each slot can attend to variable-size or more distant image regions.
After cross-attention, slots are pooled by a mean operation and fed to the policy.

\begin{table}[ht]
\caption{Scores on Crafter for agents trained on 1M environmental steps. Reported are mean scores and standard deviations of 10 random seeds. *score from~\cite{hafner2021benchmarking}.}
\vskip -0.15in
\label{tbl:crafter}
\begin{center}
\begin{small}
\begin{sc}
\begin{tabular}{lr}
\toprule
Method & Crafter Score \\
\midrule
PPO* & 4.6 $\pm$ 0.3 \\
DreamerV2* & 10.0 $\pm$ 0.2 \\
\midrule
PPO-CNN & 10.3 $\pm$ 0.6 \\
PPO-SPCNN & 11.6 $\pm$ 0.6 \\
LSTM-CNN & 10.4 $\pm$ 0.2 \\
LSTM-SPCNN & \textbf{12.1} $\pm$ 0.8 \\
OC-SA & 11.1 $\pm$ 0.7 \\
OC-CA & 10.0 $\pm$ 0.4 \\
\bottomrule
PPO-CNN (no inventory) & 6.9 $\pm$ 0.4 \\
LSTM-CNN (no inventory) & \textbf{7.7} $\pm$ 0.5 \\
\bottomrule
\end{tabular}
\end{sc}
\end{small}
\end{center}
\vskip -0.1in
\end{table}

\section{Experiments on the Crafter environment}

Here we present our findings on the original Crafter environment.
We first investigate feedforward PPO baselines both with and without inventory display.
Recurrent agents are then investigated, and we show that they outperform feedforward ones.
Finally, we evaluate asymptotic performance and show that the tuned agents can learn to unlock all but the last achievement.

\subsection{Improved baselines and hyper-parameter analysis.} 
By tuning a few hyper-parameters, we find that the simple PPO-CNN baseline outperforms the model-based DreamerV2~\cite{hafner2020mastering} (Table~\ref{tbl:crafter}).
Surprisingly, even though the PPO-SPCNN model has 134M parameters, PPO is not only able to train it, but  often outperforms other methods (Tables~\ref{tbl:crafter} and~\ref{tbl:crafterood-appearance}).
This holds for both feedforward (PPO-SPCNN) and the recurrent (LSTM-SPCNN) agents.
PPO hyper-parameters we searched over are shown in Table~\ref{tbl:ppo-hps}.
For most hyper-parameters, in the experiments we used the default values (shown in italic) that were tuned for Atari in previous work; the ones we changed are bolded.
For a detailed hyper-parameter analysis we refer to Appendix~\ref{sec:appendix:net-configs}.

The best performing agent in the Crafter Benchmark paper \cite{hafner2021benchmarking} was the DreamerV2 agent \cite{hafner2020mastering}.
Therefore, we also tried tuning its hyper-parameters: the actor entropy scale, the discount factor, the transition, and the entropy loss scales (recommended in Table B.1 of the DreamerV2 paper \cite{hafner2020mastering}), as well as the batch size, KL loss scale, reward loss scale and the discount $\lambda$.
However, our hyper-parameter search did not improve on results in \cite{hafner2021benchmarking} (see Table~\ref{tbl:dreamer-hps}).

\begin{table}[t!]
\caption{
PPO hyper-parameters we searched over.
Final used values are bolded, unless default values are used (shown in italic).
}
\vskip 0.15in
\label{tbl:ppo-hps}
\begin{center}
\begin{sc}
\begin{tabular}{ll}
\toprule
Hyper-parameter & Sweep Values \\
\midrule
Learning rate      & [0.001, 0.0005, \textit{0.0003}, 0.0001, 0.00005]  \\
Batch size         & [\textit{64}, \textbf{128}, 256]  \\
Number of rollouts & [1024, \textit{2048}, \textbf{4096}, 8196, 16384]  \\
Number of epochs   & [3, \textbf{4}, 6, 7, \textit{10}]  \\
Discount factor    & [0.8, 0.9, \textbf{0.95}, 0.97, \textit{0.99}]  \\
GAE $\lambda$      & [0.5, \textbf{0.65}, 0.75, 0.85, \textit{0.95}]  \\
Clip Range         & [0.1, \textit{0.2}, 0.3]  \\
Max Gradient Norm  & [0.1, 0.3, \textit{0.5}, 1.0]  \\
\bottomrule
\end{tabular}
\end{sc}
\end{center}
\vskip -0.1in
\end{table}

\begin{table}[t!]
\caption{DreamerV2 hyper-parameters we searched over.}
\vskip 0.15in
\label{tbl:dreamer-hps}
\begin{center}
\begin{sc}
\begin{tabular}{ll}
\toprule
Hyper-parameter & Sweep Values \\
\midrule
Batch size          & $16, 32, 64$  \\
Discount factor     & $[0.9, 0.95, 0.99, 0.999]$  \\
Actor entropy scale & $[0.003, 0.001, 0.0003, 0.0001, 0.00003]$  \\
KL loss scale       & $[0.1, 0.3, 1, 3]$  \\
Reward loss scale   & $[0.5, 1, 2]$  \\
Discount loss scale & $[0.5, 1, 2]$  \\
KL balance          & $[0.5, 0.8, 1, 2]$  \\
Discount $\lambda$  & $[0.8, 0.9, 0.95]$  \\
\bottomrule
\end{tabular}
\end{sc}
\end{center}
\vskip -0.2in
\end{table}

\subsection{Agents use inventory display as a ``scratchpad.''}
Although purely feedforward, PPO-SPCNN can unlock almost all achievements (19/22 for 1M steps and 21/22 achievements when trained for 20M steps, \cref{subsec:asymptotic}).
This requires the agent to remember unlocked achievements (e.g., I have an iron pick-axe now, need to mine a diamond next). 
However, the fact that the memory-less feedforward models unlocked most achievements leads us to hypothesize that it uses the inventory display as a ``scratchpad.''
We evaluate this by removing the inventory display, thus making the environment much less observable.
Here the agent really needs memory to know when to drink or whether it has already crafted an item or not. 
The results in Table~\ref{tbl:crafter} support our hypothesis, e.g., compare PPO-CNN with and without  the inventory display (`PPO-CNN (no inventory)`).

\subsection{Recurrent improve over feedforward agents, even without inventory information.}
The agent's reliance on the inventory display as a ``scratchpad'' led us to investigate whether recurrent agents could improve over feedforward ones by memorizing actions, maps, unlocked achievements and crafted tools in its inventory.
With the observable inventory, LSTM-CNN does not improve upon the feedforward variant PPO-CNN (10.3 vs. 10.4 in Table~\ref{tbl:crafter}).
However, when the inventory is not observable, LSTM-CNN outperforms PPO-CNN (6.9 vs. 7.7).
This might indicate that the recurrent agents learn to store achievements in memory, although not perfectly, as we observe a drop from the case with the observable inventory.
On the other hand, LSTM-SPCNN outperforms its feedforward variant PPO-SPCNN in both in-distribution and OOD settings (Table~\ref{tbl:crafterood-appearance}).
Here the inventory is observed, so the largest benefit must arise from the agent's ability to memorize the partially observed map layout.

\begin{table*}[ht!]
\caption{
Scores on CrafterOODapp (mean and standard deviations over 10 random seeds) for agents trained for 1M environmental steps. 
Each setting has two rows, denoting training (e.g. $O_{1-4}:25\%$) and evaluation ($O_1:0\%,O_{2-4}:33.3\%$) scores.
}
\vskip -0.15in
\label{tbl:crafterood-appearance}
\begin{center}
\begin{small}
\begin{sc}
\begin{tabular}{llllllll}
\toprule
Train/Eval Dist                  & PPO-CNN          & PPO-SPCNN        & LSTM-CNN         & LSTM-SPCNN                & OC-SA            & OC-CA \\
\midrule
Training: \quad \ $O_1:100\%$    & 10.4 $\pm$ 0.6   & 11.4 $\pm$ 0.5   & 10.5 $\pm$ 0.5   & 12.3 $\pm$ 0.4   & 11.0 $\pm$ 0.5   & 10.1 $\pm$ 0.6  \\
Evaluation: $O_1:100\%$          & 10.3 $\pm$ 0.6   & 11.6 $\pm$ 0.6   & 10.4 $\pm$ 0.2   & \textbf{12.1} $\pm$ 0.8   & 11.1 $\pm$ 0.7   & 10.0 $\pm$ 0.4  \\
\midrule
$O_{1-4}:25\%$                   & 9.2  $\pm$ 0.5   & 10.7 $\pm$ 0.6   & 10.7 $\pm$ 0.6   & 11.5 $\pm$ 0.4   & 9.7  $\pm$ 1.1   & 9.9 $\pm$ 0.6  \\
$O_1:0\%, \ \ O_{2-4}:33.3\%$    & 9.2  $\pm$ 0.7   & 11.0 $\pm$ 1.1   & 11.0 $\pm$ 1.0   & \textbf{11.6} $\pm$ 0.6   & 9.7  $\pm$ 1.2   & 9.2 $\pm$ 0.7  \\
\midrule
$O_1:52\%, O_{2-4}:16\%$         & 9.9 $\pm$ 0.5    & 11.1 $\pm$ 0.7   & 11.5 $\pm$ 1.4   & 11.1 $\pm$ 0.5   & 9.6  $\pm$ 0.8   & 9.4 $\pm$ 0.9  \\
$O_1:0\%, \ \ O_{2-4}:33.3\%$    & 10.0 $\pm$ 0.7   & 11.2 $\pm$ 1.1   & \textbf{11.4} $\pm$ 1.6   & 11.0 $\pm$ 0.5   & 10.6 $\pm$ 0.9   & 9.9 $\pm$ 0.9  \\
\midrule
$O_1:76\%, O_{2-4}:8\%$          & 9.9  $\pm$ 0.4   & 11.5 $\pm$ 0.6   & 10.7 $\pm$ 1.0   & 11.6 $\pm$ 0.6   & 9.8 $\pm$ 0.8    & 11.3 $\pm$ 0.5  \\
$O_1:0\%, \ \ O_{2-4}:33.3\%$    & 9.2  $\pm$ 0.6   & 10.5 $\pm$ 0.8   & 10.4 $\pm$ 1.0   & \textbf{10.7} $\pm$ 0.7   & 9.2 $\pm$ 0.8    & 10.5 $\pm$ 0.7  \\
\midrule
$O_1:88\%, O_{2-4}:4\%$          & 10.1 $\pm$ 0.6   & 12.2 $\pm$ 0.8   & 11.5 $\pm$ 1.4   & 11.3 $\pm$ 0.4   & 10.5 $\pm$ 1    & 11.2 $\pm$ 0.9  \\
$O_1:0\%, \ \ O_{2-4}:33.3\%$    & 9.1  $\pm$ 0.7   & \textbf{10.2} $\pm$ 0.7   & 10.1 $\pm$ 1.3   & 9.8  $\pm$ 0.8   & 9.4  $\pm$ 1.3  &  9.4 $\pm$ 1.0  \\
\midrule
$O_1:94\%, O_{2-4}:2\%$          & 10.9 $\pm$ 0.7   & 12.0 $\pm$ 0.8   & 11.4 $\pm$ 1.2   & 11.7 $\pm$ 0.6   & 10.5  $\pm$ 0.6  & 10.8 $\pm$ 1.1  \\
$O_1:0\%, \ \ O_{2-4}:33.3\%$    & 8.6  $\pm$ 0.7   & 9.2  $\pm$ 1.1   & 9.1  $\pm$ 1.4   & 9.8  $\pm$ 0.8   & \textbf{9.9}   $\pm$ 0.8  &  8.8 $\pm$ 0.9  \\
\midrule
$O_1:97\%, O_{2-4}:1\%$          & 10.5 $\pm$ 0.6   & 11.8 $\pm$ 0.7   & 11.9 $\pm$ 1.4   & {12.0} $\pm$ 0.3   & 10.3  $\pm$ 1.1  & 10.8 $\pm$ 0.6  \\
$O_1:0\%, \ \ O_{2-4}:33.3\%$    & 7.3  $\pm$ 0.5   & 7.7  $\pm$ 1.0   & 8.2  $\pm$ 1.0   & 8.6 $\pm$ 1.0   & \textbf{9.3}   $\pm$ 0.8  &  8.8 $\pm$ 0.8  \\
\midrule
$O_1:100\%, O_{2-4}:0\%$         & 10.5 $\pm$ 0.6   & 11.8 $\pm$ 0.6   & 10.7 $\pm$ 0.2   & {11.9} $\pm$ 0.8   & 11.1  $\pm$ 1.3  & 10.7 $\pm$ 0.6  \\
$O_1:0\%, \ \ \ \ O_{2-4}:33.3\%$& 7.3  $\pm$ 0.5   & 7.7  $\pm$ 0.9   & 5.8  $\pm$ 0.2   & 6.8  $\pm$ 1.0   & \textbf{8.0}   $\pm$ 1.2  &  7.6 $\pm$ 0.5  \\
\bottomrule
\end{tabular}
\end{sc}
\end{small}
\end{center}
\vskip -0.1in
\end{table*}

\subsection{Asymptotic Performance}
\label{subsec:asymptotic}

\begin{table}[t]
\caption{
Asymptotic performance of PPO-CNN and PPO-SPCNN on the Crafter environment up to 20M environmental interactions. 
}
\vskip -0.15in
\label{tbl:crafter-asymptotic}
\begin{center}
\begin{small}
\begin{sc}
\begin{tabular}{lcccccc}
\toprule
Method    & 1M   & 2M   & 5M   & 10M  & 15M  & 20M  \\
\midrule
PPO-CNN   & 10.3 & 15.9 & 18.3 & 27.1 & 28.3 & 29.1 \\
PPO-SPCNN & 11.6 & 16.5 & 19.4 & 27.8 & 29.6 & \textbf{30.5} \\
\bottomrule
\end{tabular}
\end{sc}
\end{small}
\end{center}
\vskip -0.1in
\end{table}

As the Crafter benchmark is designed to accelerate RL research and facilitate short research cycles, the default evaluation protocol is to train agents up to 1M steps.
To better understand the asymptotic performance, we train PPO-CNN and PPO-SPCNN agents for 20M steps.
The resulting mean Crafter scores over ten seeds are shown in \cref{tbl:crafter-asymptotic}.
Other agents do not improve significantly over these two in the asymptotic regime.
For readability, we omit standard deviations (they are all below 0.6, decreasing to 0.2 at 20M steps due to the geometric mean averaging in the score).
Firstly, PPO-SPCNN consistently outperforms the PPO-CNN baseline.
Moreover, the PPO-SPCNN agent learns to unlock all but one achievement (the diamond) at 5M steps (though this is not a regular event).
We suspect that the difficulty in mining the diamond is its rarity, implying that better exploration strategies are needed.
Unlike humans, the agents have no prior notion that a diamond is valuable.
This challenge seems ideal for investigating novel curiosity-based agents \cite{schmidhuber1990making,schmidhuber1991curious,schmidhuber2013powerplay}.

\section{OOD Generalization Experiments}

In this section, we evaluate agents on new CrafterOODapp and CrafterOODnum environments, showing limited generalization of baseline agents and improved generalization of object-centric ones.

\begin{table*}[t]
\caption{
Scores on CrafterOODnum (mean and standard deviations over 10 random seeds) for agents trained for 1M environmental steps. 
Each setting has two rows, denoting scores on the training and evaluation environments.
}
\vskip -0.15in
\label{tbl:crafterood-numbers}
\begin{center}
\begin{small}
\begin{sc}
\begin{tabular}{llllllll}
\toprule
Train/Eval Dist & PPO-CNN    & PPO-SPCNN        & LSTM-CNN         & LSTM-SPCNN      & OC-SA            & OC-CA \\
\midrule
Train: easy (x2) & 12.4 $\pm$ 2   & 13.7 $\pm$ 1.2   & 14.2 $\pm$ 0.8   & 13.3 $\pm$ 1.0   & 15.0 $\pm$ 1.8   & 13.7 $\pm$ 1.8  \\
Eval: default    & 10.4 $\pm$ 1.7 & 12.1 $\pm$ 1.0   & 11.5 $\pm$ 0.7   & 11.4 $\pm$ 0.9   & \textbf{13.0} $\pm$ 1.3   & 11.7 $\pm$ 1.7  \\
\midrule
easy (x4) & 13.1 $\pm$ 2.2  & 14.5 $\pm$ 1.8  & 15.1 $\pm$ 1.2   & 15.0 $\pm$ 0.7   & 18.6 $\pm$ 2.3   & 13.7 $\pm$ 1.9  \\
default   & 8.8  $\pm$ 1.0  & 9.8  $\pm$ 1.7  & 9.1  $\pm$ 0.9   & 9.8  $\pm$ 0.9   & \textbf{12.8} $\pm$ 1.7   & 8.8  $\pm$ 0.9  \\
\midrule
mix (x4)  & 13.4 $\pm$ 1.7  & 13.9 $\pm$ 2.1  & 13.0 $\pm$ 0.7   & 14.7 $\pm$ 1.6   & 15.5 $\pm$ 2.0   & 14.7 $\pm$ 2.3  \\
default   & 9.2  $\pm$ 0.6  & 10.2 $\pm$ 1.7  & 9.0  $\pm$ 0.9   & 10.5 $\pm$ 0.9   & \textbf{10.6} $\pm$ 1.7   & 9.4 $\pm$ 0.9  \\
\midrule
default   & 10.2 $\pm$ 0.4  & 11.5 $\pm$ 0.5  & 10.7 $\pm$ 0.6   & 12.1 $\pm$ 0.6   & 11.3 $\pm$ 0.4   & 10.2 $\pm$ 0.7  \\
mix (x4)  & 11.2 $\pm$ 1.1  & 12.3 $\pm$ 1.0  & 11.6 $\pm$ 0.9   & 12.2 $\pm$ 1.3   & \textbf{12.6} $\pm$ 1.0   & 10.3 $\pm$ 0.9  \\
\midrule
default   & 10.3 $\pm$ 0.4  & 11.2 $\pm$ 0.6  & 10.9 $\pm$ 0.7   & 12.0 $\pm$ 0.4   & 11.1 $\pm$ 0.5   & 10.1 $\pm$ 0.6  \\
easy (x2) & 10.9 $\pm$ 1.4  & \textbf{13.7} $\pm$ 1.3   & 12.8 $\pm$ 0.9   & 12.9 $\pm$ 1.4   & 11.4 $\pm$ 1.2   &  10.1 $\pm$ 0.7  \\
\midrule
default   & 10.3 $\pm$ 0.7  & 11.2 $\pm$ 0.3  & 10.6 $\pm$ 0.3   & 11.8 $\pm$ 0.7   & 11.5 $\pm$ 0.7   & 10.1 $\pm$ 0.6  \\
easy (x4) & 11.3 $\pm$ 0.8  & 12.5 $\pm$ 0.9  & 12.8 $\pm$ 1.3   & 11.9 $\pm$ 1.0   & \textbf{12.9} $\pm$ 0.8   & 9.9 $\pm$ 1.2  \\
\midrule
easy (x2) & 11.7 $\pm$ 1.3  & 13.3 $\pm$ 0.4  & 13.7 $\pm$ 1.4  & 13.4 $\pm$ 0.2   & 15.4  $\pm$ 1.2  & 12.8 $\pm$ 1.6  \\
hard (x2) & 8.0  $\pm$ 0.5  & 8.1  $\pm$ 1.7  & 7.6  $\pm$ 0.7  & 9.2  $\pm$ 0.8   & \textbf{10.5}  $\pm$ 1.5  &  7.1 $\pm$ 0.8  \\
\midrule
easy (x4) & 14.6 $\pm$ 2.2  & 15.4 $\pm$ 1.6  & 15.5 $\pm$ 1.2  & 15.3 $\pm$ 1.0   & 17.8  $\pm$ 1.5  & 15.2 $\pm$ 1.8  \\
hard (x4) & 3.0  $\pm$ 0.4  & 3.3  $\pm$ 0.4  & 3.0  $\pm$ 0.3  & 3.4  $\pm$ 0.4   & \textbf{4.9}   $\pm$ 0.6  &  4.2 $\pm$ 0.3  \\
\midrule
Train: average & 12.0 $\pm$ 1.4 & 13.1 $\pm$ 1.1 & 13.0 $\pm$ 0.8 & 13.4 $\pm$ 0.8 & 14.5 $\pm$ 1.3  & 12.6 $\pm$ 1.4  \\
Eval: average  & 9.1 $\pm$ 0.9  & 10.2 $\pm$ 1.1 & 9.7 $\pm$ 0.9  & 9.1 $\pm$  1.0 & \textbf{11.1} $\pm$ 1.1 &  8.9 $\pm$ 0.9  \\
\bottomrule
\end{tabular}
\end{sc}
\end{small}
\end{center}
\vskip -0.1in
\end{table*}

\begin{table}[t]
\caption{
CrafterOODnum environment object numbers.\\
Default is the original Crafter environment.
Easy environments have (x2 or x4) more resources and (x2 or x4) fewer enemies.
In hard environments we decrease the number of resources and increase the number of enemies.
Mix environment contains four times more resources and enemies.
}
\vskip -0.15in
\label{tbl:crafteroodnum}
\begin{center}
\begin{small}
\begin{sc}
\begin{tabular}{lccccc}
\toprule
Env. & Tree & Coal & Cow & Zombie & Skeleton \\
\midrule
Easy (x4)    & 764 & 206  & 100 & 3  & 2.5 \\
Easy (x2)    & 380 & 102  &  46 & 6  & 4.5 \\
Default      & 189 &  50  &  26 & 15 & 9.5 \\
Hard (x2)    &  95 &  27  &  13 & 33 & 19  \\
Hard (x4)    &  52 & 12.5 &   6 & 60 & 38  \\
Mix (all x4) & 764 & 206  & 100 & 60 & 38  \\
\bottomrule
\end{tabular}
\end{sc}
\end{small}
\end{center}
\vskip -0.1in
\end{table}

\subsection{CrafterOODapp - out-of-distribution object appearance.}

In the CrafterOODapp environment, we generate a collection of increasingly complex adaptation scenarios.
Environments contain up to four object variants for trees, cows, iron, stones, zombies, and skeletons, the objects the agent interacts with most frequently (\cref{fig:textures-ood}).
The adaptation scenarios differ in object distributions in training and evaluation environments.
For example, the in-distribution training and evaluation corresponds to observing only the first object variant $O_1$ (first two rows in Table~\ref{tbl:crafterood-appearance}).

Starting from training with uniformly distributed objects ($O_{1-4}=25\%$) we make generalization progressively more difficult by skewing the training distribution towards the first object.
In this way, we create environments that contain the first object ($O_1$) in $52\%$ and all the others in  $16\%$ of the time ($O_1:52\%, O_{2-4}:16\%$ case in \cref{tbl:crafterood-appearance}), then environments that contain the first object in $76\%$, then $88\%$ percent of the time, etc.
The evaluation environment always contains only the last three objects (we refer to them as the ``evaluation'' objects) uniformly distributed (each observed in $33.3\%$ of the time).
Up to the point of observing the evaluation objects in $16\%$ of the time the agents generalize fairly well (\cref{tbl:crafterood-appearance}, but see also \cref{fig:crafterood-heatmap} in Appendix~\ref{sec:appendix:crafterood-heatmap}).
Decreasing the percentage of evaluation objects in the training environment further, the performance of all agents consistently drops.
Finally, in the especially difficult zero-shot generalization scenario, when trained only on the first object ($O_1=100\%$), the agent relies on pure chance or interacting with objects that do not change (e.g., water, iron).
The agent's failure to generalize is expected: to perform well on unseen evaluation objects, it must perform fast adaptation and systematic generalization in terms of composing the previously obtained knowledge about the observed objects and the inferred representations of the newly introduced evaluation objects.

Motivated by this observation, we introduce object-centric agents, which should (at least in principle) learn better representations by decomposing the input images into sets of objects and their distances via positional encodings.
Our experiments show that the object-centric agents OC-SA and OC-CA (the last two columns in \cref{tbl:crafterood-appearance}) match the vanilla PPO agents on in-distribution and easy OOD generalization.
In the most difficult OOD generalization cases though (the last three pairs of rows in \cref{tbl:crafterood-appearance}), OC-SA and OC-CA generalize better compared to PPO-CNN and PPO-SPCNN (9.9 vs 8.6 and 9.2 for $O_1=94\%$, 9.3 vs 7.3 and 7.7 for $O_1=97\%$, and 8.0 vs 7.3 and 7.7 for $O_1=100\%$).
Additionally, object-centric agents are interpretable and potentially easier to build on in future work.
We visualize their attention maps in Section~\ref{subsec:oc-attn}.
The observed generalization gap makes environments with large differences in object distributions (requiring zero-shot adaptation) fruitful for developing and evaluating novel fast-adaptation \cite{Schmidhuber:92selfref, Schmidhuber:93selfrefann, Schmidhuber:93selfreficann, Schmidhuber:93selfrefieee, irie2022modern} and meta-learning \cite{schmidhuber1987evolutionary} agents, such as the ones based on Fast Weight Programmers \cite{Schmidhuber:91fastweights, schmidhuber1992learning, schmidhuber1993reducing, schlag2021linear, irie2021going}.

\subsection{CrafterOODnum - out-of-distribution object numbers.}

To test generalization on different numbers of objects, we introduce the CrafterOODnum environment.
We vary the numbers of objects the agent interacts with most frequently: trees, coal, cows, zombies, and skeletons.
These variants are particularly challenging as the agent might face never-before-seen situations such as fighting against more enemies or surviving in environments with resources scarcer compared to what it was trained on.
In \cref{tbl:crafterood-numbers} we show the numbers of objects in the respective environments.
In \cref{tbl:crafterood-numbers}, `Default` is the Crafter environment introduced in \cite{hafner2021benchmarking}.
In `Easy` environments, we increase (double or quadruple) the number of objects representing resources and decrease the number of enemies.
`Hard` environments consist of fewer resources and more enemies.
In the `mix` environment, we quadruple the numbers of all objects---making it on the one hand easier in terms of resources, but on the other hand harder in terms of enemies.

For each pair of training and evaluation environments in \cref{tbl:crafterood-numbers} the agents with the best generalization score are bolded.
The object-centric agent OC-SA outperforms all models in all but one case.
Note how the performance decreases when transferring to harder environments (Easy to Default) and increases when transferring to easier environments (Default to Easy), confirming the intended design difficulty level.
The performance drop is largest when transferring from `Easy (x4)` to `Hard (x4)` environments where the evaluation environment contains 16x fewer resources and 16x more enemies than the training environment.
Finally, the object-centric OC-SA agent achieves the best generalization on average across all the environments (the last row in \cref{tbl:crafterood-numbers}).

\begin{figure*}[t]
	\centering
	\includegraphics[width=\linewidth]{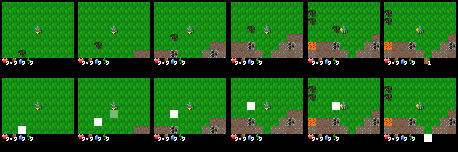}
	\caption{\it 
	An agent collecting resources.
	Top row: input images. Bottom row: attention visualized by its intensity.  Episode steps are plotted horizontally.}
	\label{fig:attn2}
\end{figure*}

\section{Object-centric Agents Analysis}

In this section, we first perform an ablation study of object-centric agents, inspecting various design choices.
Afterwards, we visualize attention patterns of object-centric agents, and find their policies to be interpretable and to match our gameplay intuitions.

\begin{table}[ht!]
\caption{OC-CA Agents Network Ablation.}
\vskip -0.15in
\label{tbl:cross-attn-ablation}
\begin{center}
\begin{sc}
\begin{tabular}{lr}
\toprule
Variant & Crafter Score \\
\midrule
\textbf{OC-CA}                & \textbf{10.0 $\pm$ 0.4} \\
OC-CA + Residual MLP & 7.3  $\pm$ 0.4 \\
OC-CA + LayerNorm    & 4.1  $\pm$ 0.3 \\
OC-CA + Residual MLP + LayerNorm & 3.1  $\pm$ 0.6 \\
OC-CA + Slot Competition     & 7.1  $\pm$ 0.3 \\
\midrule
OC-CA, Number of Slots=1 & 8.2 $\pm$ 0.9 \\
OC-CA, Number of Slots=2 & 7.8 $\pm$ 0.4 \\
OC-CA, Number of Slots=4 & 8.7 $\pm$ 0.7 \\
\textbf{CA, Number of Slots=8} & \textbf{10.0 $\pm$ 0.4} \\
OC-CA, Number of Slots=16 & 9.1 $\pm$ 0.9 \\
\midrule
OC-CA, Number of Heads=1 & 6.6 $\pm$ 0.5 \\
OC-CA, Number of Heads=2 & 7.3 $\pm$ 0.9 \\
OC-CA, Number of Heads=4 & 8.0 $\pm$ 0.6 \\
\textbf{CA, Number of Heads=8} & \textbf{10.0 $\pm$ 0.4} \\
CA, Number of Heads=16 & 7.2 $\pm$ 0.8 \\
\midrule
OC-CA, Patch Size=1, Stride=1    & 6.9  $\pm$ 0.9 \\
OC-CA, Patch Size=8, Stride=8    & 7.2  $\pm$ 0.4 \\
OC-CA, Patch Size=12, Stride=8   & 8.7  $\pm$ 0.7 \\
OC-CA, Patch Size=12, Stride=12   & 8.9  $\pm$ 0.7 \\
OC-CA, Patch Size=16, Stride=8   & 6.2  $\pm$ 0.5 \\
\textbf{OC-CA, Patch Size=16, Stride=16}   & \textbf{10.0  $\pm$ 0.4} \\
\bottomrule
\end{tabular}
\end{sc}
\end{center}
\vskip -0.1in
\end{table}

\begin{figure*}[t]
	\centering
	\includegraphics[width=\linewidth]{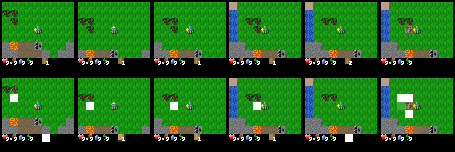}
	\caption{\it 
	An agent building a table.
	Top row: input images. Bottom row: attention visualized by its intensity.  Episode steps are plotted horizontally. }
	\label{fig:attn3}
\end{figure*}

\begin{figure*}[t]
	\centering
	\includegraphics[width=\linewidth]{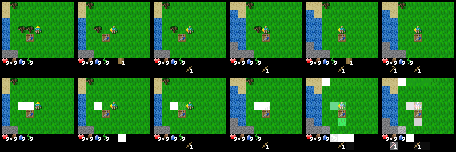}
	\caption{\it 
	An agent crafting weapons.
	Top row: input images. Bottom row: attention visualized by its intensity.  Episode steps are plotted horizontally. }
	\label{fig:attn4}
\end{figure*}

\begin{figure*}[t]
	\centering
	\includegraphics[width=0.98\linewidth]{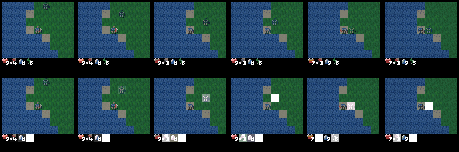}
	\caption{\it 
	An agent defending against enemies. Top row: input images. Bottom row: attention visualized by its intensity.  Episode steps are plotted horizontally. 
	}
	\label{fig:attn5}
\end{figure*}

\begin{figure*}[ht!]
	\centering
	\includegraphics[width=0.98\linewidth]{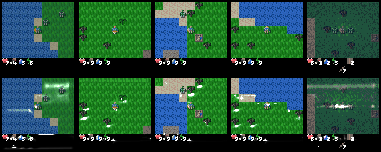}
	\caption{\it 
	Learned attention over the whole input image instead of a fixed patch grid.
	Top row: input images. Bottom row: attention visualized by its intensity.  
	}
	\label{fig:attn-learned}
\end{figure*}

\subsection{Object-centric Agents Ablation.}
\label{subsec:oc-perf}
To the best of our knowledge, this is the first time cross-attention-based methods (e.g. SlotAttention- and Perceiver-like methods) are used in an open world survival game.
In this section, we perform an ablation analysis, compare our agent to the standard architectural choices from the literature and show that these vanilla agents underperform.

The ablation study results are shown in Table~\ref{tbl:cross-attn-ablation}.
Firstly, cross-attention-based methods typically employ LayerNorm~\cite{ba2016layer} and residual MLPs. 
In Table~\ref{tbl:cross-attn-ablation} we can see that the variant using these ``CA + Residual MLP + LayerNorm'' underperforms.
Including any of these modules individually (``CA + Residual MLP'' and ``CA + LayerNorm'') does not improve performance.
Moreover, competition over slots (a softmax over queries) ``CA + Slot Competition'' akin to SlotAttention also hurts the downstream performance.
This led us to converge to an architecture that uses a single cross-attention over the input image, with no latent self-attention, no layer normalization, no residual connections, and no slot-wise competition.

We found the optimal number of heads and slots to be eight.
We speculate that the higher number of slots lets each attend to different objects in the input, and the higher number of heads allows for operation specialization for each of the heads.
Finally, we observe that a larger patch size (with the appropriate stride) improves performance.
On the other hand, the agents with learned object representations ``CA, Patch Size=1, Stride=1'' do not perform as well as the attention with larger patches.
Although in theory more powerful as it can attend to varying object sizes, this variant needs first to \textit{learn} which pixels belong to each object.
In Crafter, objects are of equal size, so the patches can be chosen to correspond to those.
We suspect the attention over pixels would better generalize to objects of different sizes.

\subsection{OC-SA Agents Visualization and Interpretability.}
\label{subsec:oc-attn}

Here we visualize the CLS token attention in OC-SA.
Visualizations in \cref{fig:attn2,fig:attn3,fig:attn4,fig:attn5} stem from a single agent and are representative of most episodes of a trained agent.
In \cref{fig:attn2}, we can see an agent collecting resources. 
In the first frame (the leftmost figure) the agent notices a tree nearby and then focuses consistently on it and collects it. 
In the final frame, the agent attends to the newly collected wood resource.

The episode is continued in Figure~\ref{fig:attn3}, in which the agent builds a table.
After collecting one tree, the agent needs to collect another one to build a table. 
It spots more trees in its vicinity and goes towards them, never losing sight of the tree and the position where it wants to craft a table.
It collects another tree and attends again to the wood resources.
Determining that it has enough wood (two pieces), it builds a table to facilitate crafting of further tools and weapons.

Once it has built the table, the agent can use it to craft tools (weapons), but it first needs to collect more resources. 
In Figure~\ref{fig:attn4} it collects the two nearby trees and uses them to build a wooden sword and a wooden pickaxe (shown in the inventory). 
The sequence is reminiscent of what a human player would do: collect one tree, craft a sword, collect another tree and craft a pickaxe.
It can use pickaxe to collect stones or coal or to defend against enemies.

In \cref{fig:attn5}, an agent defends against a zombie during the night (the reason why the frame color is darker). 
Initially, the agent does not attend to the zombie, probably because it is still far away. 
As the zombie gets closer, it attends more and more to it and finally defeats it.

\subsection{OC-CA Agents Visualization and Interpretability.}

In Figure~\ref{fig:attn-learned} we visualize the learned attention patterns when each slot can attend to any set of pixels in the input image, compared to only a predefined set of patches from a fixed grid.
We find that the learned attention attends to the salient objects in the scene: zombies (1st, 3rd, and 5th columns), trees (2nd, 3rd, and 4th columns), resources (water overall), and the inventory. 
We observe more numerous but smaller attention patterns than the patch-based attention that is typically focused on two to five patches.
Although in theory more powerful, in practice the learned attention models underperform the patch-based ones.
We suspect this is because the learned attention agents first need to find out what an object actually is, whereas patch-based ones have a more ``guided'' learning process.
We speculate that greater gains of learned attention would occur in scenarios with objects of varying sizes, in which case a fixed grid would be a too rigid representation.

\section{Related Work}

RL benchmarks have played a crucial role in developing and evaluating novel (deep) RL algorithms \cite{bellemare13arcade,brockman2016openai,kempka2016vizdoom,beattie2016deepmind,tassa2018deepmind,juliani2018unity}.
An especially important role has been played by video games including arcade games \cite{bellemare13arcade}, racing environments \cite{wymann2000torcs}, first-person shooters \cite{kempka2016vizdoom}, strategy games \cite{synnaeve2016torchcraft,vinyals2017starcraft}, and open world games \cite{guss2019neurips,guss2021minerl}.
For an extensive survey of deep (reinforcement) learning for video game playing see \cite{justesen2019deep}.
Open world games have recently received special attention, e.g., Minecraft  \cite{johnson2016malmo,guss2019neurips,guss2021minerl}.

However, these environments come with their drawbacks.
For example, Minecraft is too complex to be solved from scratch by current methods \cite{milani2020retrospective}.
The recently published method for mining a diamond \cite{baker2022video} uses human-annotated data to bootstrap learning via behavioral cloning. 
It is also unclear by what metric agents should be evaluated.
On the other hand, Atari requires large amounts of computation: training an agent with five random seeds on each game for 200M steps requires over 2000 GPU days \cite{castro2018dopamine,hessel2018rainbow, hafner2021benchmarking}, which hinders fast iteration cycles.
Most importantly, many Atari games are nearly deterministic, so the agents can approximately memorize action sequences and are not required to generalize to new situations \cite{machado2018revisiting}.

Environments like ProcGen \cite{cobbe2020leveraging} evaluate the generalization capabilities of agents through procedural level generation, which requires substantial compute  \cite{hafner2021benchmarking}.
Chan et al. \cite{chan2022zipfian} recently investigated how RL algorithms perform in environments where feature distribution is not uniform but Zipfian, similar to the one of objects encountered in the real world, or in our CrafterOODapp environments. 
Generalization benchmarks based on DeepMind's Control Suite \cite{tassa2018deepmind,stone2021distracting,hansen2021generalization,grigsby2020measuring,zhang2018natural}
test the agent's robustness to background variations  but not to variations of environmental object configurations, e.g., visual appearance, number of objects, or object compositions.
For a thorough recent survey on generalization in RL, see \cite{kirk2021survey}.

The Crafter game addresses these shortcomings.
Crafter is an open world survival game for RL research whose dynamics are inspired by the popular game Minecraft.
The benchmark is designed to address existing research challenges, such as strong generalization via procedural generation, deep exploration via achievements conditioned on one another, learning from high-dimensional image observations and sparse rewards that require long-term reasoning and credit assignment.
It facilitates evaluation by combining semantically meaningful achievements and fast iteration speed.
Our newly introduced CrafterOODapp and CrafterOODnum inherit all these benefits and additionally test some agent's generalization ability and robustness in presence of unseen objects, setting the stage for the development of fast-adaptation or meta-learning agents.

Research on RL neural network-based policies that generalize to OOD environments is an important research area currently tackled from several directions \cite{kirk2021survey}.
One potential explanation of the poor generalization of neural networks is that they cannot dynamically and flexibly bind the information distributed throughout the network \cite{greff2020binding}. 
This is also known as the binding problem  \cite{von1994correlation, roskies1999binding}.
Recent work addresses this through object-centric neural networks that learn (discrete) object representations from raw visual input, to support efficient learning and generalization to novel scenarios and behaviors.
Object-centric methods have successfully been used for OOD generalization in both supervised and unsupervised learning \cite{greff2015binding,greff2016tagger,greff2017neural,van2018relational,eslami2016attend,kosiorek2018sequential,stanic2019rsqair,burgess2019monet,greff2019multi,engelcke2019genesis,locatello2020object,stanic2020hierarchical,creswell2021unsupervised,kipf2021conditional} and in RL \cite{watters2019cobra,veerapaneni2020entity,kipf2020Contrastive,carvalho2021feature}.
To the best of our knowledge, this is the first time they are used in open world RL survival games.

Architecturally closest to our work are methods that learn a set of vectors (slots) by attending to the input via cross-attention.
This idea was introduced in SetTransformer \cite{lee2019set} and successfully used for object detection \cite{carion2020end}, unsupervised learning of objects \cite{locatello2020object,kipf2021conditional}, learning permutation-invariant agents \cite{tang2020neuroevolution,tang2021sensory}, and general perception modules \cite{jaegle2021perceiver,jaegle2021perceiverio,alayrac2022flamingo}.
Most similar to our object-centric agents are the AttentionAgent~\cite{tang2020neuroevolution} and the SensoryNeuron \cite{tang2021sensory}.
The former uses neuroevolution to optimize an architecture with a hard attention bottleneck, resulting in a network that only receives a fraction of the visual input and generalizes to unseen backgrounds. 
The latter further improves AttentionAgent's robustness to permuted orderings of its inputs.
Our agents also share similarities with attention-based modular neural networks \cite{santoro2018relational,goyal2019recurrent,mott2019towards,carvalho2021feature}, learning a (modular) representation by attending over the input.
However, unlike our object-centric work, prior work did not consider object-centric agents for OOD generalization in procedurally generated open world games.

\section{Conclusion and Discussion}

Challenging benchmarks are essential for research on new reinforcement learning methods. 
However, they should be simple enough to allow for convenient systematic analysis and fast iteration cycles.
Tests against strong baselines are crucial.

Our paper offers contributions to all of the above.
We report important observations on the Crafter environment.
Our analysis shows that PPO-based agents trained for 20M steps can unlock all but the last achievement.
However, this impressive score still falls short of the human score \cite{hafner2021benchmarking}, indicating that Crafter remains an open research challenge, especially for sample-efficient RL agents.

We introduce CrafterOODapp and CrafterOODnum, new environments that evaluate some agent's robustness against varying object appearances and numbers of objects.
Baseline agents fail to adapt in evaluation environments containing unseen objects, whereas our novel object-centric agents perform well.
An additional benefit of object-centric agents is their interpretability.
In the investigated environments, self-attention-based methods outperform cross-attention methods. 
Future work will confirm or disconfirm similar superiority in environments with varying object sizes.

Our OOD environments should help to study fast-adaptation and meta-learning agents.
We intend to further improve object-centric agents by incorporating decoding objectives, to extract as much signal from the environment as possible, and by exploring ways of training larger object-centric networks.
We also want to investigate whether object-centric inductive biases can enable model-based RL agents to learn better world models. Crafter and our newly introduced CrafterOOD are good candidates for exploring all of these directions, offering control of environmental configurations, excellent visual inspection, and fast iteration cycles.
To facilitate future research in this area, we will publish the code for our agents and environments.

\section*{Acknowledgements}
We thank Kazuki Irie, Danijar Hafner, Shixiang Shane Gu and Yingtao Tian for their useful comments and suggestions on an earlier version of this paper. 
This research was partially funded by the ERC Advanced grant no: 742870, project AlgoRNN.

\bibliography{biblio}
\bibliographystyle{IEEEtran}

\newpage
\appendices

\onecolumn
\section{OOD environment Objects}

\begin{figure*}[h!]
\centering
\providecommand{\radius}{1ex}
\providecommand{\size}{0.12\textwidth}
\imageWithCorners{\radius}{\size}{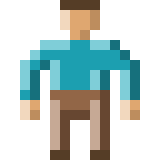}\hfill%
\imageWithCorners{\radius}{\size}{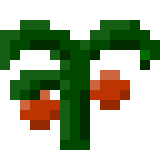}\hfill%
\imageWithCorners{\radius}{\size}{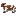}\hfill%
\imageWithCorners{\radius}{\size}{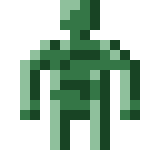}\hfill%
\imageWithCorners{\radius}{\size}{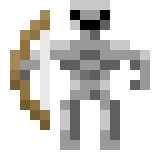}\hfill%
\imageWithCorners{\radius}{\size}{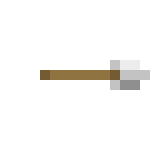} \\
\makebox[\size]{Player}\hfill%
\makebox[\size]{Plant}\hfill%
\makebox[\size]{Cow}\hfill%
\makebox[\size]{Zombie}\hfill%
\makebox[\size]{Skeleton}\hfill%
\makebox[\size]{Arrow} \\[1.5ex]
\imageWithCorners{\radius}{\size}{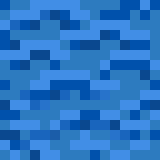}\hfill%
\imageWithCorners{\radius}{\size}{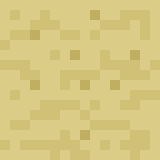}\hfill%
\imageWithCorners{\radius}{\size}{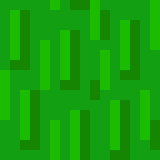}\hfill%
\imageWithCorners{\radius}{\size}{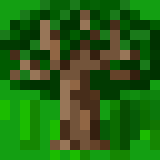}\hfill%
\imageWithCorners{\radius}{\size}{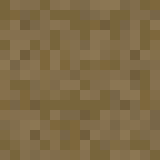}\hfill%
\imageWithCorners{\radius}{\size}{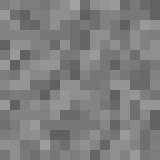} \\
\makebox[\size]{Water}\hfill%
\makebox[\size]{Sand}\hfill%
\makebox[\size]{Grass}\hfill%
\makebox[\size]{Tree}\hfill%
\makebox[\size]{Path}\hfill%
\makebox[\size]{Stone} \\[1.5ex]
\imageWithCorners{\radius}{\size}{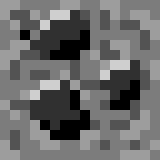}\hfill%
\imageWithCorners{\radius}{\size}{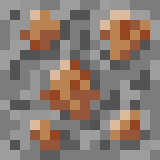}\hfill%
\imageWithCorners{\radius}{\size}{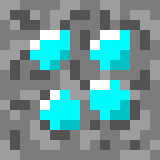}\hfill%
\imageWithCorners{\radius}{\size}{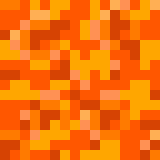}\hfill%
\imageWithCorners{\radius}{\size}{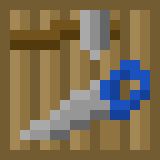}\hfill%
\imageWithCorners{\radius}{\size}{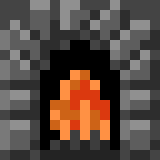} \\
\makebox[\size]{Coal}\hfill%
\makebox[\size]{Iron}\hfill%
\makebox[\size]{Diamond}\hfill%
\makebox[\size]{Lava}\hfill%
\makebox[\size]{Table}\hfill%
\makebox[\size]{Furnace} \\
\caption{Original Crafter objects. Figure from \cite{hafner2021benchmarking}.}
\label{fig:textures}
\end{figure*}

\begin{figure*}[h!]
\centering
\providecommand{\radius}{1ex}
\providecommand{\size}{0.12\textwidth}
\imageWithCorners{\radius}{\size}{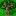}\hfill%
\imageWithCorners{\radius}{\size}{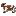}\hfill%
\imageWithCorners{\radius}{\size}{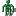}\hfill%
\imageWithCorners{\radius}{\size}{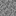}\hfill%
\imageWithCorners{\radius}{\size}{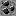}\hfill%
\imageWithCorners{\radius}{\size}{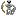} \\
\makebox[\size]{TreeV1}\hfill%
\makebox[\size]{CowV1}\hfill%
\makebox[\size]{ZombieV1}\hfill%
\makebox[\size]{StoneV1}\hfill%
\makebox[\size]{CoalV1}\hfill%
\makebox[\size]{SkeletonV1} \\[1.5ex]
\imageWithCorners{\radius}{\size}{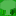}\hfill%
\imageWithCorners{\radius}{\size}{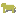}\hfill%
\imageWithCorners{\radius}{\size}{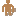}\hfill%
\imageWithCorners{\radius}{\size}{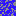}\hfill%
\imageWithCorners{\radius}{\size}{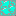}\hfill%
\imageWithCorners{\radius}{\size}{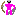} \\
\makebox[\size]{TreeV2}\hfill%
\makebox[\size]{CowV2}\hfill%
\makebox[\size]{ZombieV2}\hfill%
\makebox[\size]{StoneV2}\hfill%
\makebox[\size]{CoalV2}\hfill%
\makebox[\size]{SkeletonV2} \\[1.5ex]
\imageWithCorners{\radius}{\size}{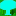}\hfill%
\imageWithCorners{\radius}{\size}{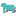}\hfill%
\imageWithCorners{\radius}{\size}{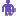}\hfill%
\imageWithCorners{\radius}{\size}{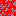}\hfill%
\imageWithCorners{\radius}{\size}{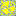}\hfill%
\imageWithCorners{\radius}{\size}{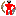} \\
\makebox[\size]{TreeV3}\hfill%
\makebox[\size]{CowV3}\hfill%
\makebox[\size]{ZombieV3}\hfill%
\makebox[\size]{StoneV3}\hfill%
\makebox[\size]{CoalV3}\hfill%
\makebox[\size]{SkeletonV3} \\[1.5ex]
\imageWithCorners{\radius}{\size}{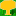}\hfill%
\imageWithCorners{\radius}{\size}{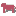}\hfill%
\imageWithCorners{\radius}{\size}{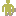}\hfill%
\imageWithCorners{\radius}{\size}{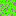}\hfill%
\imageWithCorners{\radius}{\size}{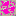}\hfill%
\imageWithCorners{\radius}{\size}{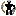} \\
\makebox[\size]{TreeV4}\hfill%
\makebox[\size]{CowV4}\hfill%
\makebox[\size]{ZombieV4}\hfill%
\makebox[\size]{StoneV4}\hfill%
\makebox[\size]{CoalV4}\hfill%
\makebox[\size]{SkeletonV4} \\
\caption{In CrafterOODapp there are four variants of objects for trees, cows, zombies, stone coal and skeletons.}
\label{fig:textures-ood}
\end{figure*}

\newpage
\twocolumn
\section{Network Configurations}
\label{sec:appendix:net-configs}

\begin{table}[!htb]
    \centering
    \caption{PPO-CNN: baseline agent with the CNN from DQN~\cite{mnih2015human}.}
    \vskip 0.15in
    \begin{tabular}{l}
        \toprule
        \textbf{Feature Extractor} \\
            \midrule 
            $8\times 8$ conv, $32$ ReLU units, stride $4$ \\
            $4\times 4$ conv, $64$ ReLU units, stride $2$\\
            $3\times 3$ conv, $64$ ReLU units, stride $1$\\
            Flatten \\
            Linear, $512$ ReLU units. \\
            \midrule 
        \textbf{Action Network} \\
            \midrule 
            Linear, $17$ units. \\
            \midrule 
        \textbf{Value Network} \\
            \midrule 
            Linear, $1$ units. \\
        \bottomrule
    \end{tabular}
    \label{tbl:ppo-cnn}
\end{table}

\begin{table}[!htb]
    \centering
    \caption{PPO-SPCNN: agent with the size-preserving CNN.}
    \vskip 0.15in
    \begin{tabular}{l}
        \toprule
        \textbf{Feature Extractor} \\
            \midrule 
            $5\times 5$ conv, $64$ ReLU units, stride $1$, padding $2$ \\
            $5\times 5$ conv, $64$ ReLU units, stride $1$, padding $2$ \\
            $5\times 5$ conv, $64$ ReLU units, stride $1$, padding $2$ \\
            $5\times 5$ conv, $64$ ReLU units, stride $1$, padding $2$ \\
            Flatten \\
            Linear, $512$ ReLU units. \\
            \midrule 
        \textbf{Action Network} \\
            \midrule 
            Linear, $17$ units. \\
            \midrule 
        \textbf{Value Network} \\
            \midrule 
            Linear, $1$ units. \\
        \bottomrule
    \end{tabular}
    \label{tbl:ppo-spcnn}
\end{table}

\begin{table}[!htb]
    \centering
    \caption{OC-SA: agent with the self-attention module.}
    \vskip 0.15in
    \begin{tabular}{l}
        \toprule
        \textbf{Feature Extractor} \\
            \midrule 
            $5\times 5$ conv, $64$ ReLU units, stride $1$, padding $2$ \\
            $5\times 5$ conv, $64$ ReLU units, stride $1$, padding $2$ \\
            $5\times 5$ conv, $64$ ReLU units, stride $1$, padding $2$ \\
            $5\times 5$ conv, $64$ ReLU units, stride $1$, padding $2$ \\
            Split into $8\times 8$ patches and flatten the patch grid \\
            \midrule 
        \textbf{Self-Attention Network} \\
            \midrule 
            Learned CLS token of $256$ size. \\
            Slot-wise projection: Linear, $256$ units. \\
            // Self-Attention layer 1: \\
            Query map: Linear, $256$ units. \\
            Key map: Linear, $256$ units. \\
            Values map: Linear, $256$ units. \\
            // Self-Attention layer 2: \\
            Query map: Linear, $256$ units. \\
            Key map: Linear, $256$ units. \\
            Values map: Linear, $256$ units. \\
            \midrule 
        \textbf{Action Network} \\
            \midrule 
            Linear, $17$ units. \\
            \midrule 
        \textbf{Value Network} \\
            \midrule 
            Linear, $1$ units. \\
        \bottomrule
    \end{tabular}
    \label{tbl:OC-SA}
\end{table}

\begin{table}[!htb]
    \centering
    \caption{OC-CA: agent with the cross-attention module.}
    \vskip 0.15in
    \begin{tabular}{l}
        \toprule
        \textbf{Feature Extractor} \\
            \midrule 
            $5\times 5$ conv, $64$ ReLU units, stride $1$, padding $2$ \\
            $5\times 5$ conv, $64$ ReLU units, stride $1$, padding $2$ \\
            $5\times 5$ conv, $64$ ReLU units, stride $1$, padding $2$ \\
            $5\times 5$ conv, $64$ ReLU units, stride $1$, padding $2$ \\
            Split into $16\times 16$ patches and flatten the patch grid \\
            \midrule 
        \textbf{Self-Attention Network} \\
            \midrule 
            Learned slots $8\times 256$ size. \\
            Query map: Linear, $256$ units. \\
            Key map: Linear, $256$ units. \\
            Values map: Linear, $256$ units. \\
            \midrule 
        \textbf{Action Network} \\
            \midrule 
            Linear, $17$ units. \\
            \midrule 
        \textbf{Value Network} \\
            \midrule 
            Linear, $1$ units. \\
        \bottomrule
    \end{tabular}
    \label{tbl:OC-CA}
\end{table}

\begin{table}[!htb]
    \centering
    \caption{LSTM-CNN: recurrent agent with an LSTM and a CNN from DQN~\cite{mnih2015human}.}
    \vskip 0.15in
    \begin{tabular}{l}
        \toprule
        \textbf{Feature Extractor} \\
            \midrule 
            $8\times 8$ conv, $32$ ReLU units, stride $4$ \\
            $4\times 4$ conv, $64$ ReLU units, stride $2$\\
            $3\times 3$ conv, $64$ ReLU units, stride $1$\\
            Flatten \\
            Linear, $512$ ReLU units. \\
            \midrule 
        \textbf{Action Network} \\
            \midrule 
            LSTM, $256$ units. \\
            Linear, $17$ units. \\
            \midrule 
        \textbf{Value Network} \\
            \midrule 
            LSTM, $256$ units. \\
            Linear, $1$ units. \\
        \bottomrule
    \end{tabular}
    \label{tbl:LSTM-cnn}
\end{table}

\begin{table}[!htb]
    \centering
    \caption{LSTM-SPCNN: recurrent agent with an LSTM and a size-preserving CNN.}
    \vskip 0.15in
    \begin{tabular}{l}
        \toprule
        \textbf{Feature Extractor} \\
            \midrule 
            $5\times 5$ conv, $64$ ReLU units, stride $1$, padding $2$ \\
            $5\times 5$ conv, $64$ ReLU units, stride $1$, padding $2$ \\
            $5\times 5$ conv, $64$ ReLU units, stride $1$, padding $2$ \\
            $5\times 5$ conv, $64$ ReLU units, stride $1$, padding $2$ \\
            Flatten \\
            Linear, $512$ ReLU units. \\
            \midrule 
        \textbf{Action Network} \\
            \midrule 
            LSTM, $256$ units. \\
            Linear, $17$ units. \\
            \midrule 
        \textbf{Value Network} \\
            \midrule 
            Linear, $256$ units. \\
            Linear, $1$ units. \\
        \bottomrule
    \end{tabular}
    \label{tbl:LSTM-spcnn}
\end{table}

\newpage\phantom{blabla}
\newpage
\onecolumn

\section{CrafterOODapp performance}
\label{sec:appendix:crafterood-heatmap}

\begin{figure}[h]
	\centering
	\includegraphics[width=\linewidth]{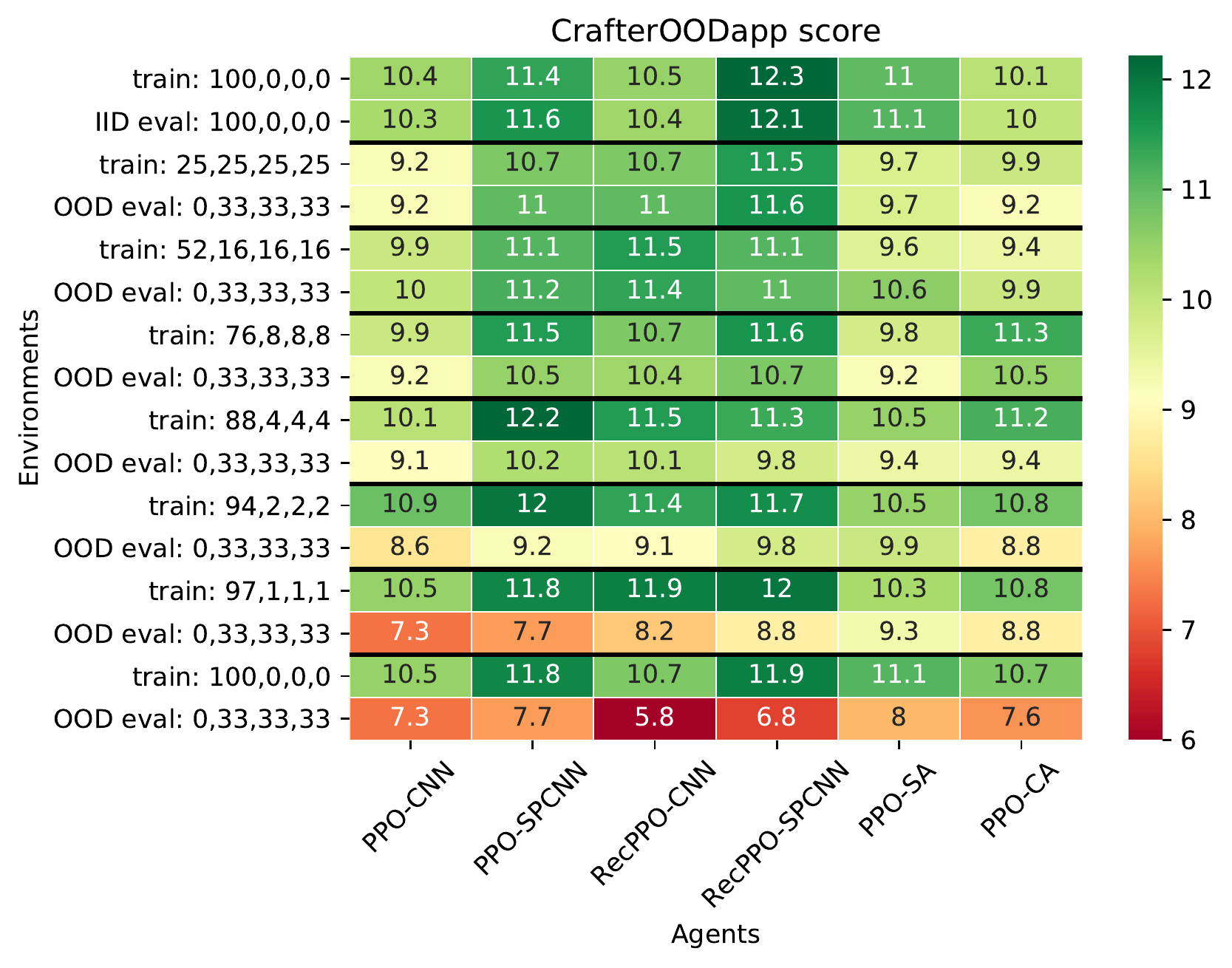}
	\caption{\it 
	Scores on CrafterOODnum for agents trained for 1M environment steps. 
	Mean over 10 random seeds are reported. For standard deviations see \cref{tbl:crafterood-appearance}.
	Each setting has two rows, denoting scores in training (e.g. $25,25,25,25$ for $O_{1-4}:25\%$) and evaluation ($0,33,33,33$, for $O_1:0\%,O_{2-4}:33.3\%$) environments.
	}
	\label{fig:crafterood-heatmap}
\end{figure}

In \cref{fig:crafterood-heatmap} we provide heatmap of the CrafterOODapp scores previously reported in \cref{tbl:crafterood-appearance}.

\section{CrafterOODnum performance}
\label{sec:appendix:crafterood-heatmap-num}

\begin{figure}[h]
	\centering
	\includegraphics[width=\linewidth]{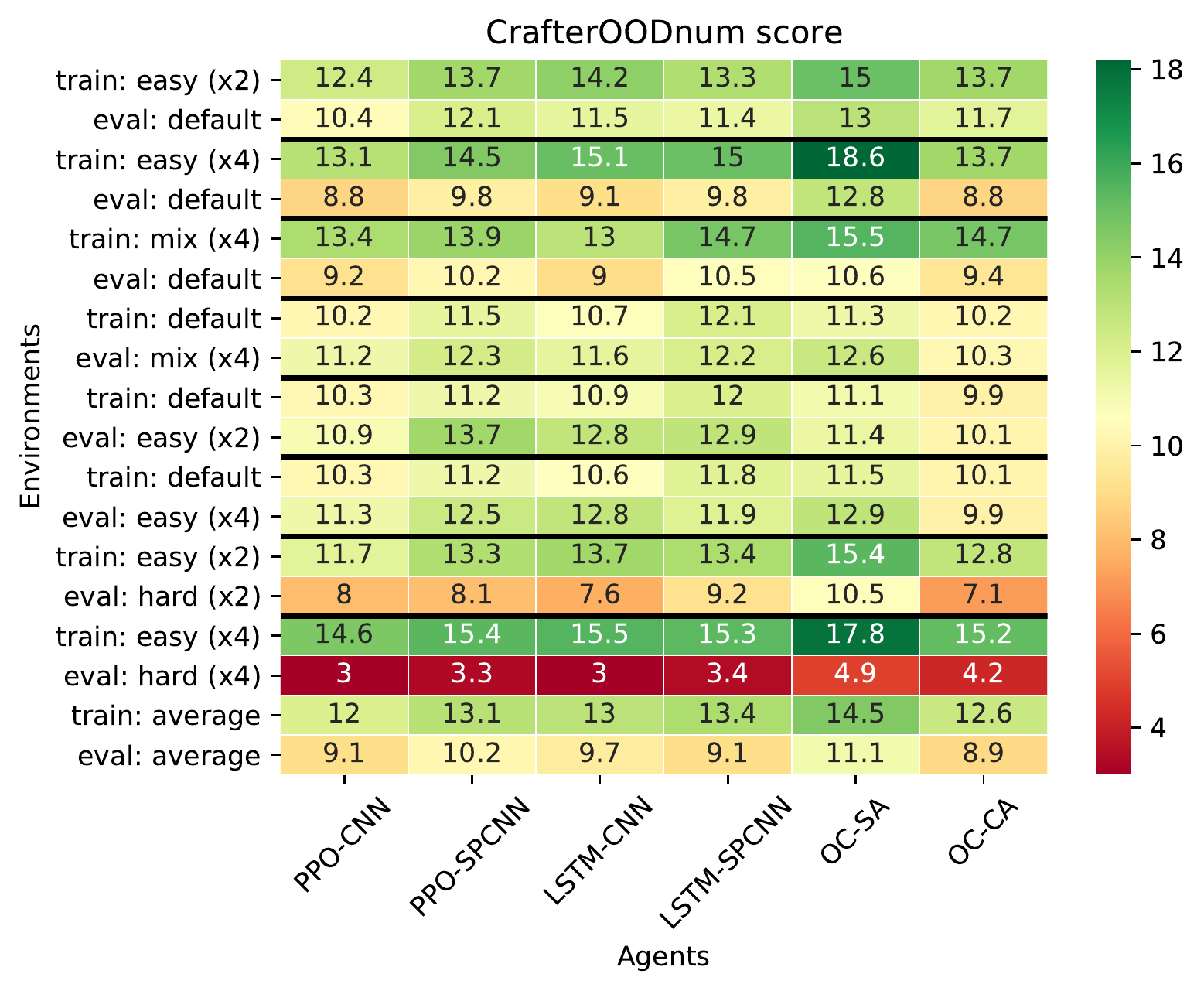}
	\caption{\it 
	Scores on CrafterOODapp for agents trained for 1M environment steps. 
	Mean over 10 random seeds are reported. For standard deviations see \cref{tbl:crafterood-numbers}.
	Each setting has two rows, denoting scores in training and evaluation environments.
	}
	\label{fig:crafterood-num-heatmap}
\end{figure}

In \cref{fig:crafterood-num-heatmap} we provide heatmap of the CrafterOODnum scores previously reported in \cref{tbl:crafterood-numbers}.

\section{Hyper-parameter heatmaps}
\label{sec:appendix:hp-heatmaps}

\begin{figure}[ht]
    \centering
	\subfloat
	{
        \includegraphics[width=0.38\linewidth]{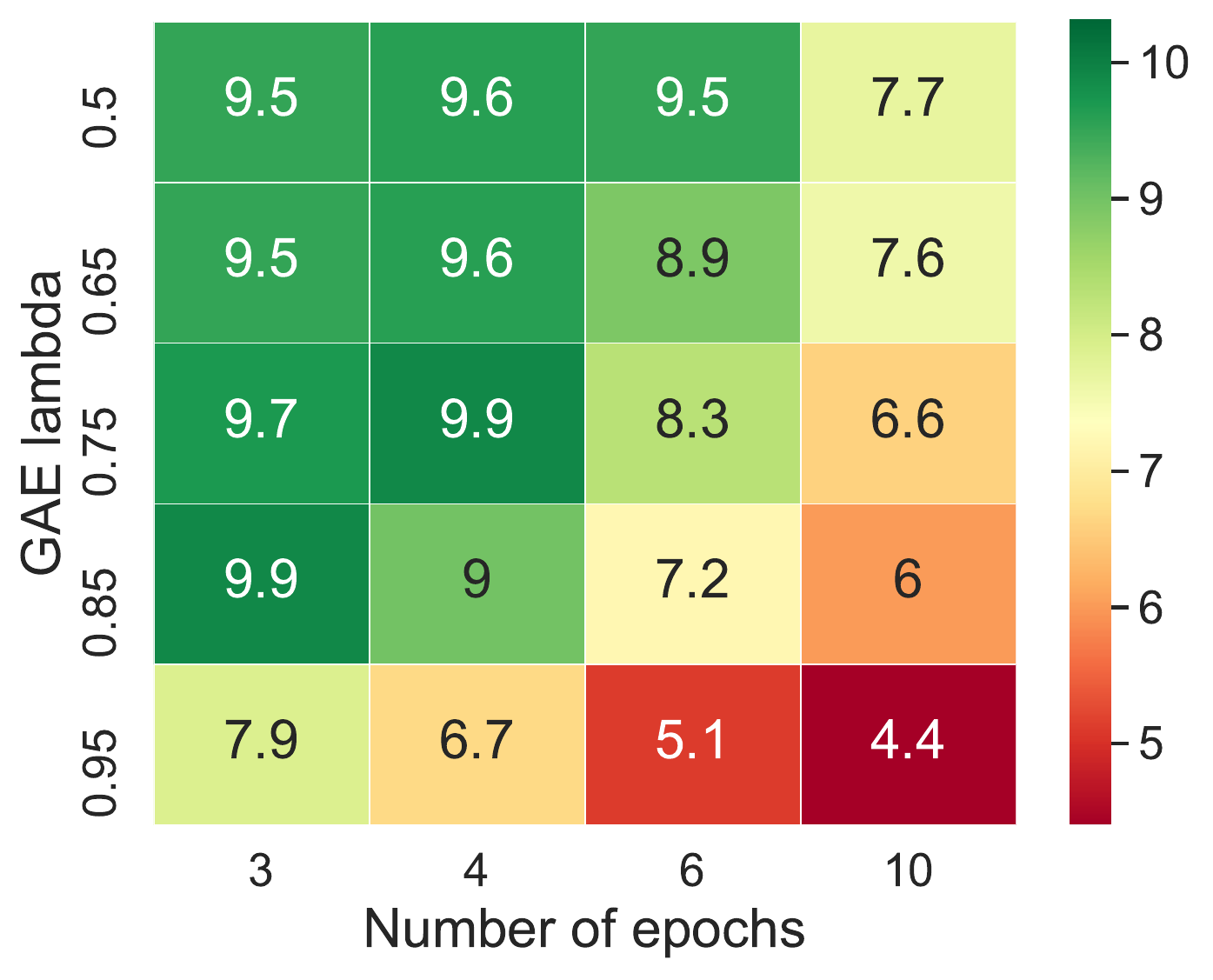}
        \label{sfig:hp-heatmap-cnndef12}
	}
	\subfloat
	{
        \includegraphics[width=0.38\linewidth]{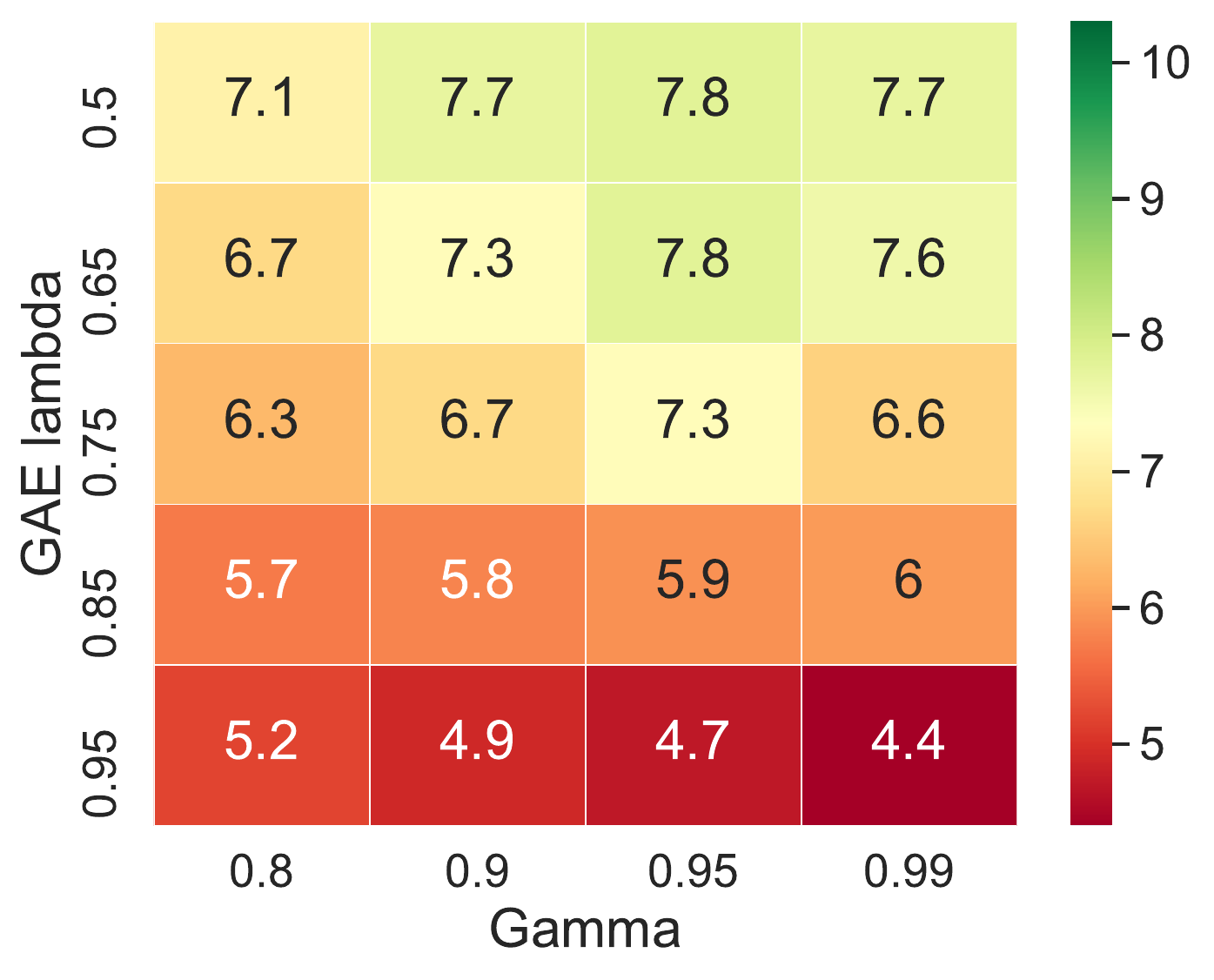}
        \label{sfig:hp-heatmap-cnndef13}
	}
\hfil
	\subfloat
	{
        \includegraphics[width=0.38\linewidth]{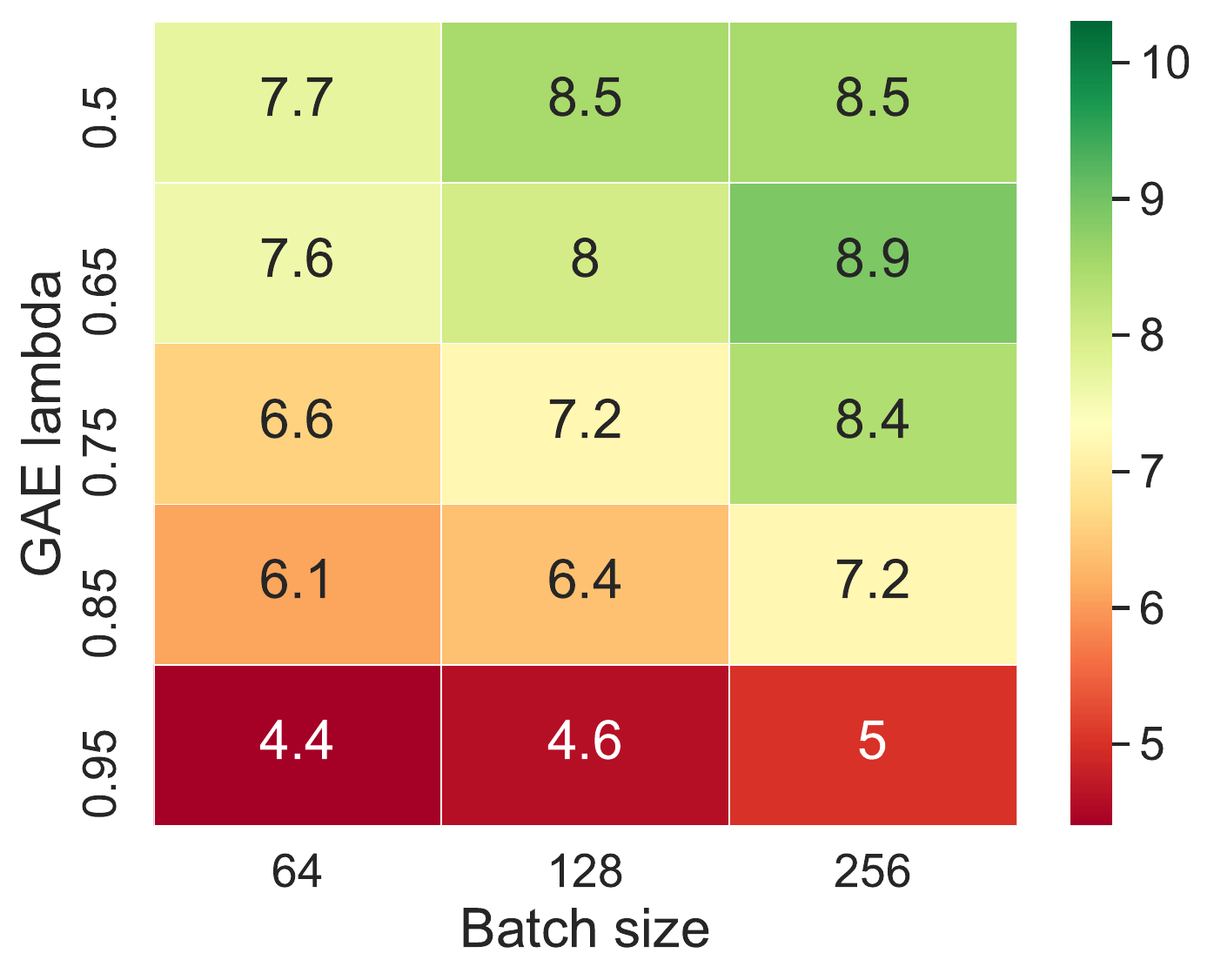}
        \label{sfig:hp-heatmap-cnndef14}
	}
	\subfloat
	{
        \includegraphics[width=0.38\linewidth]{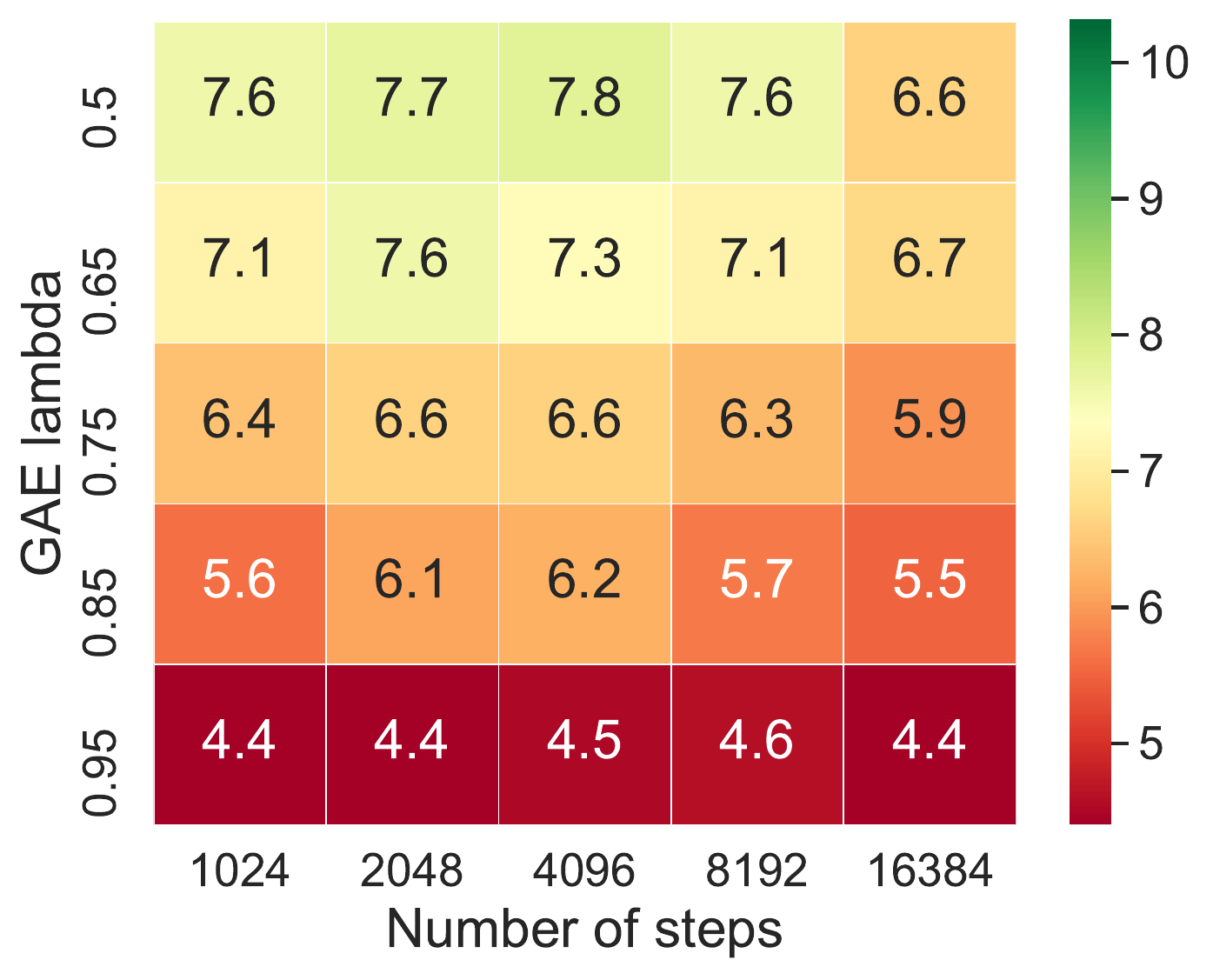}
        \label{sfig:hp-heatmap-cnndef15}
	}
\hfil
	\subfloat
	{
        \includegraphics[width=0.38\linewidth]{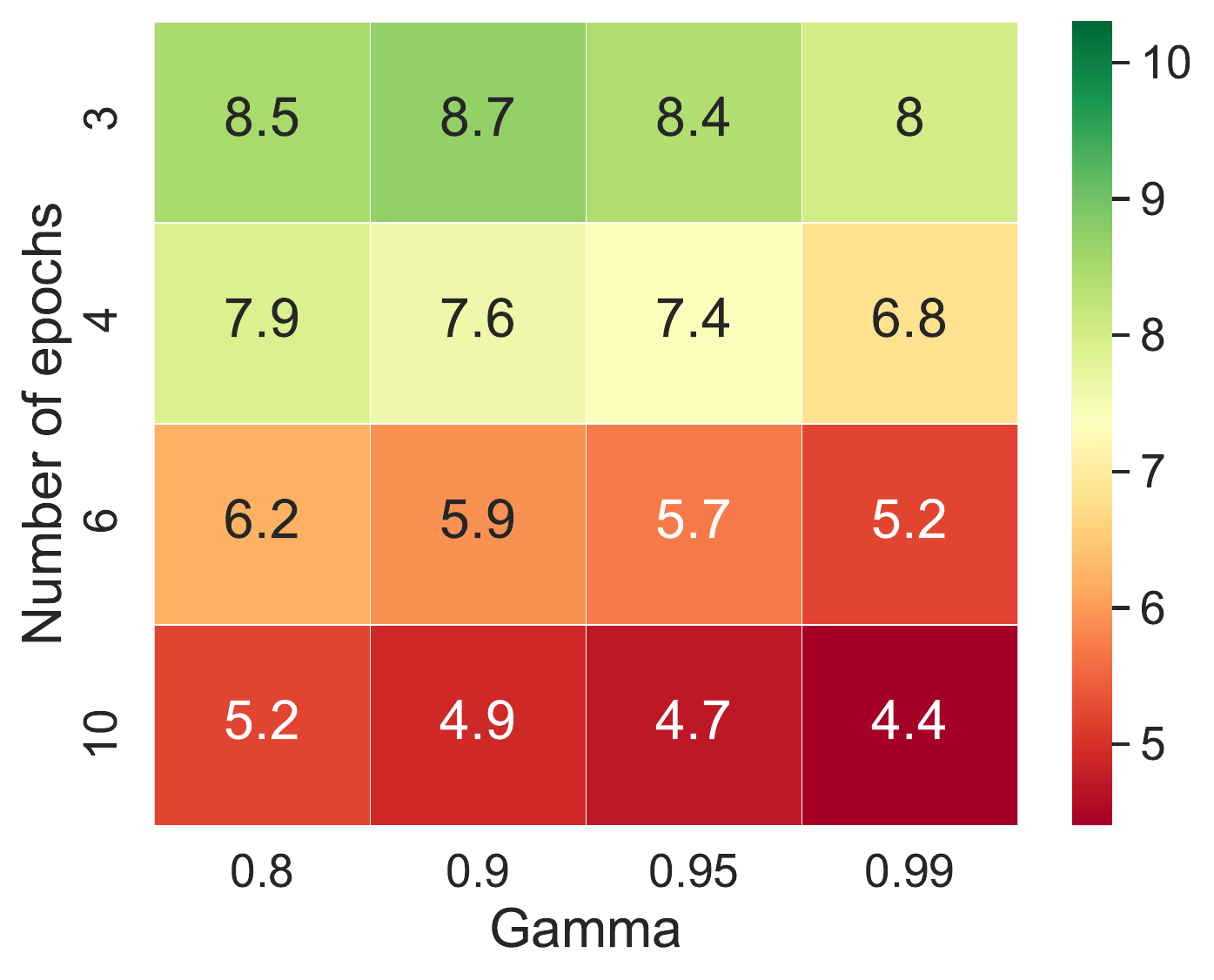}
        \label{sfig:hp-heatmap-cnndef23}
	}
	\subfloat
	{
        \includegraphics[width=0.38\linewidth]{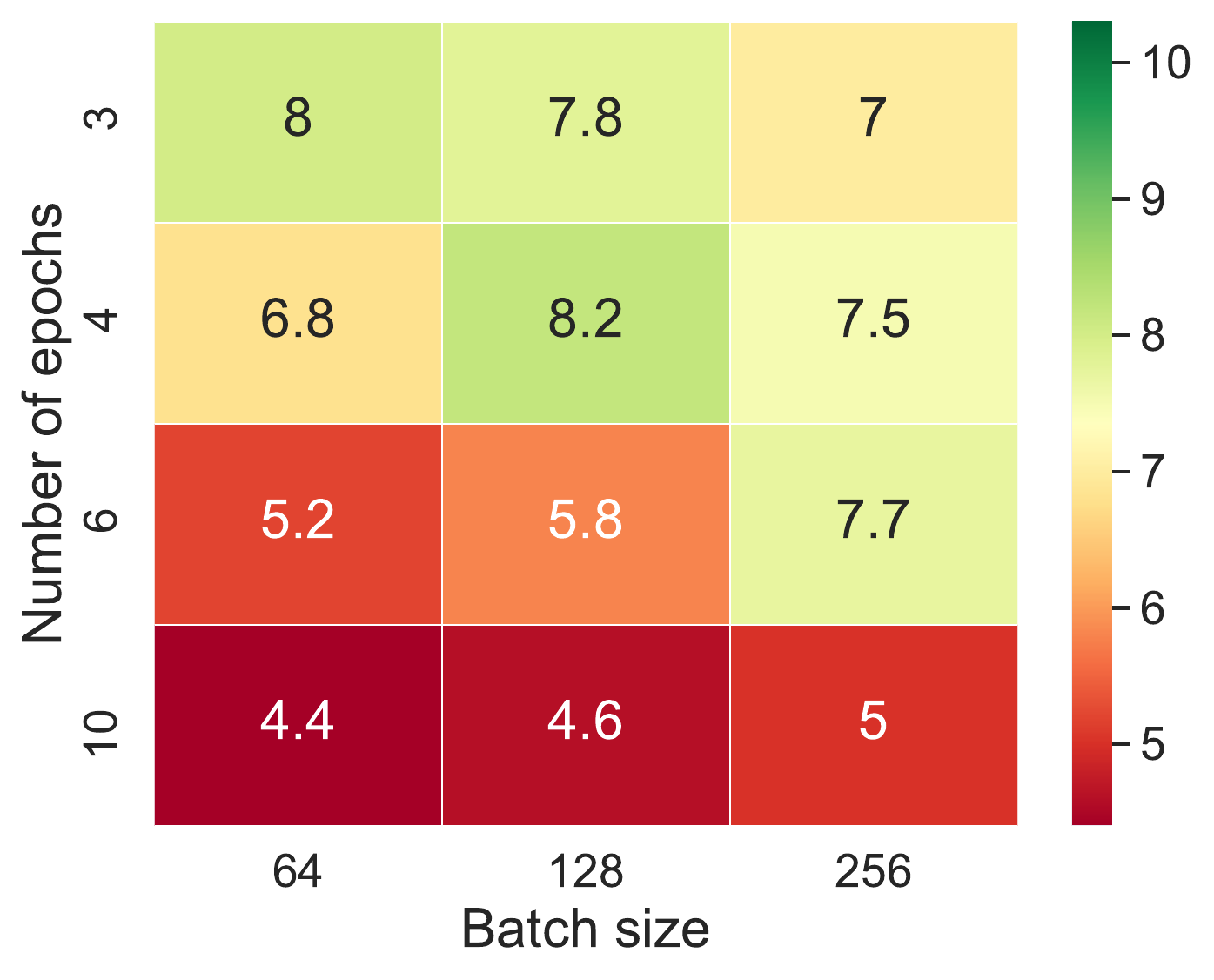}
        \label{sfig:hp-heatmap-cnndef24}
	}
\hfil
	\subfloat
	{
        \includegraphics[width=0.38\linewidth]{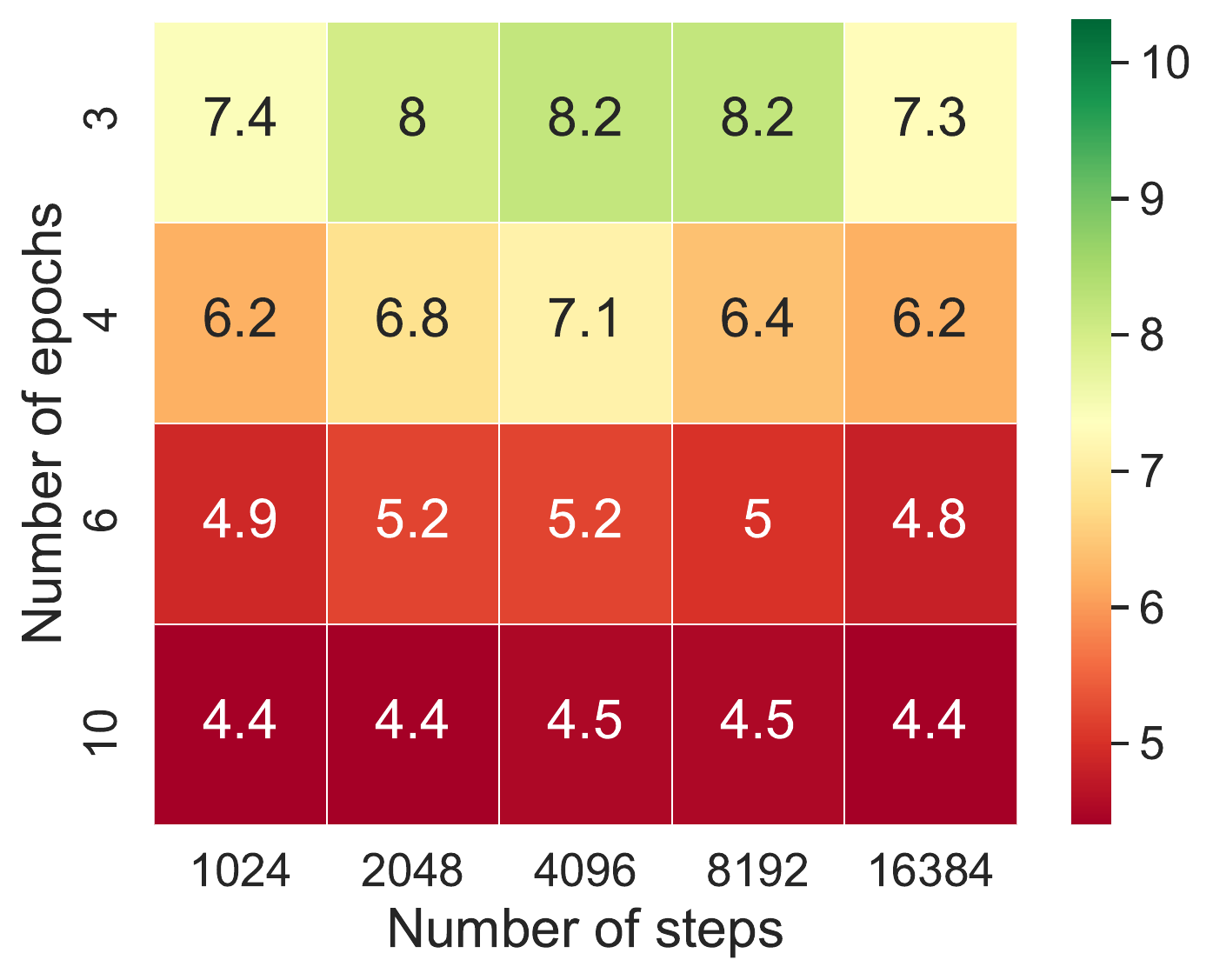}
        \label{sfig:hp-heatmap-cnndef25}
	}
	\subfloat
	{
        \includegraphics[width=0.38\linewidth]{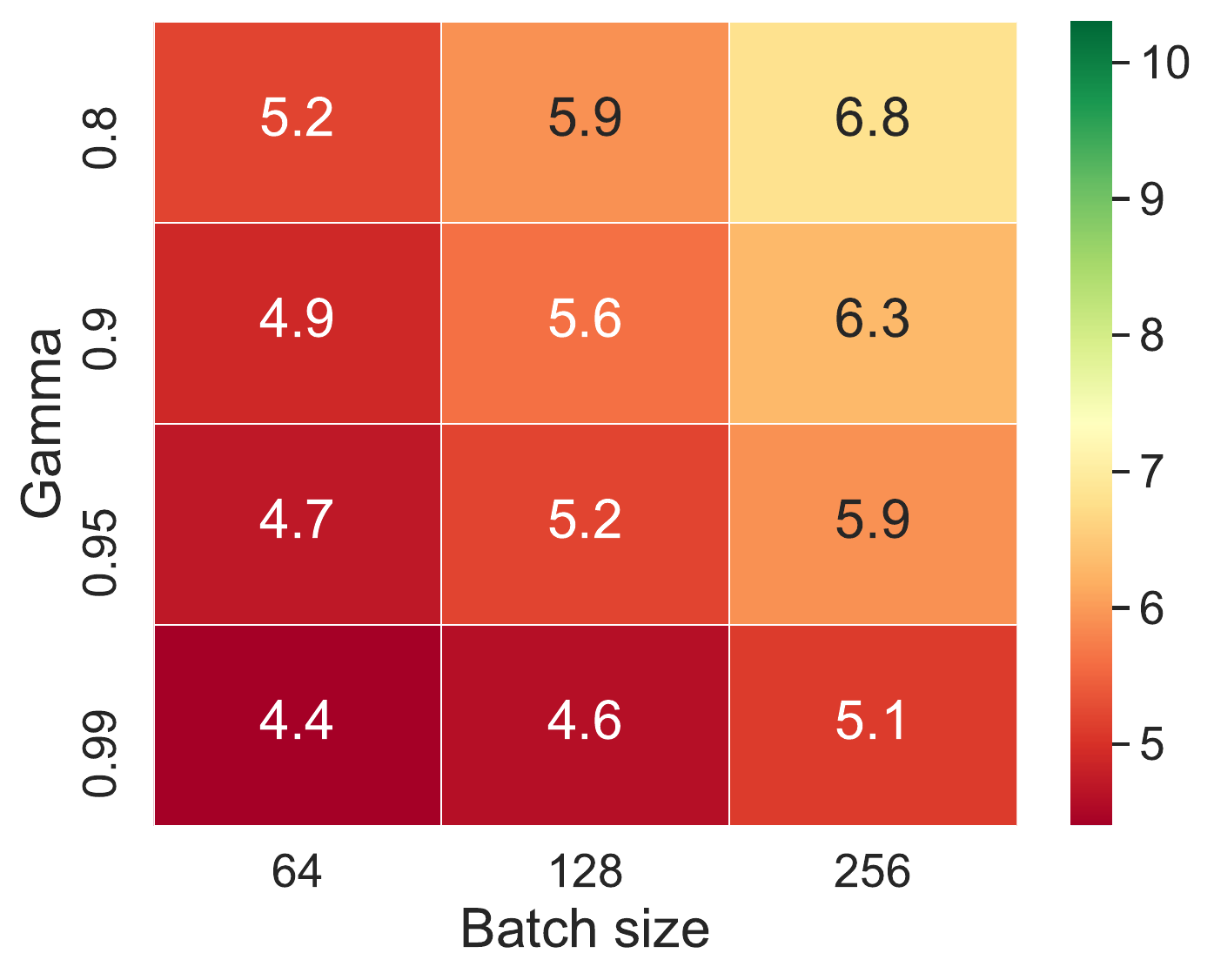}
        \label{sfig:hp-heatmap-cnndef34}
	}
\hfil
	\caption{\it 
	\centering
	Crafter scores starting from the default hyper-parameters from \cite{stable-baselines3} that were tuned for Atari: GAE lambda 0.95, number of epochs 10, gamma 0.99, batch size 64 and number of steps 1024, and then ablating individual hyper-parameter pairs.
	}
	\label{fig:hp-heatmap-cnndef}
\end{figure}

\begin{figure}[ht]
    \centering
	\subfloat
	{
        \includegraphics[width=0.38\linewidth]{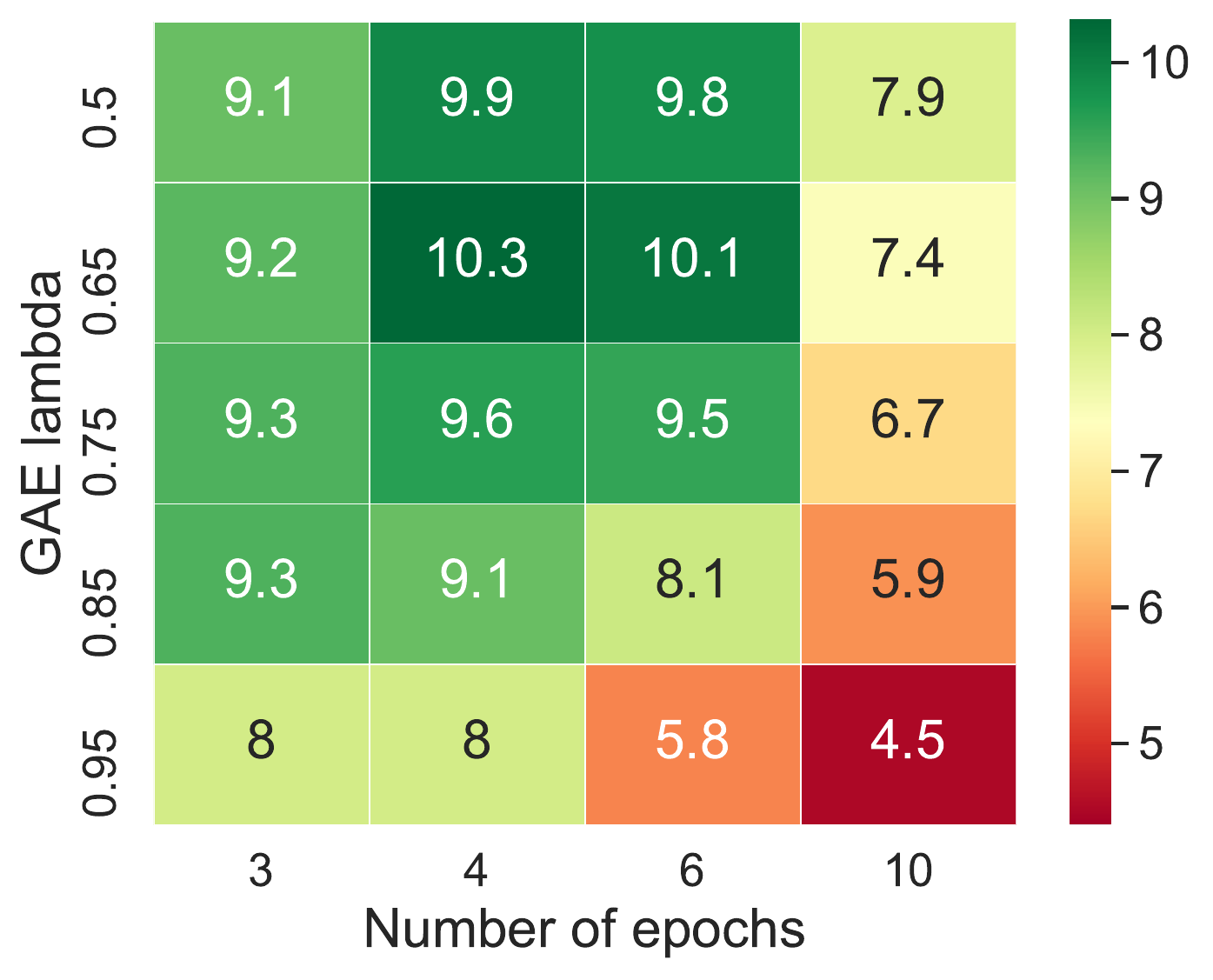}
        \label{sfig:hp-heatmap-cnnbest12}
	}
	\subfloat
	{
        \includegraphics[width=0.38\linewidth]{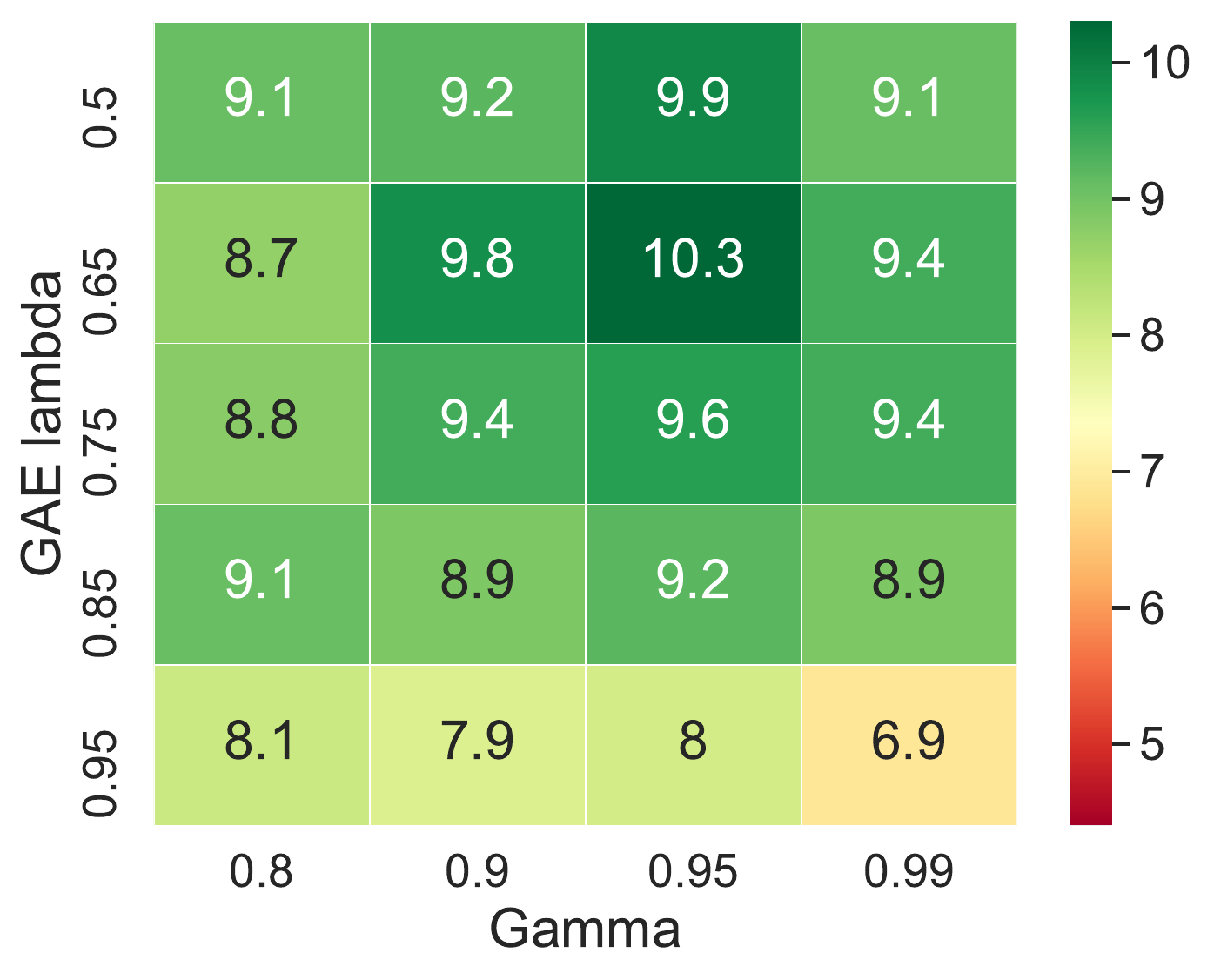}
        \label{sfig:hp-heatmap-cnnbest13}
	}
\hfil
	\subfloat
	{
        \includegraphics[width=0.38\linewidth]{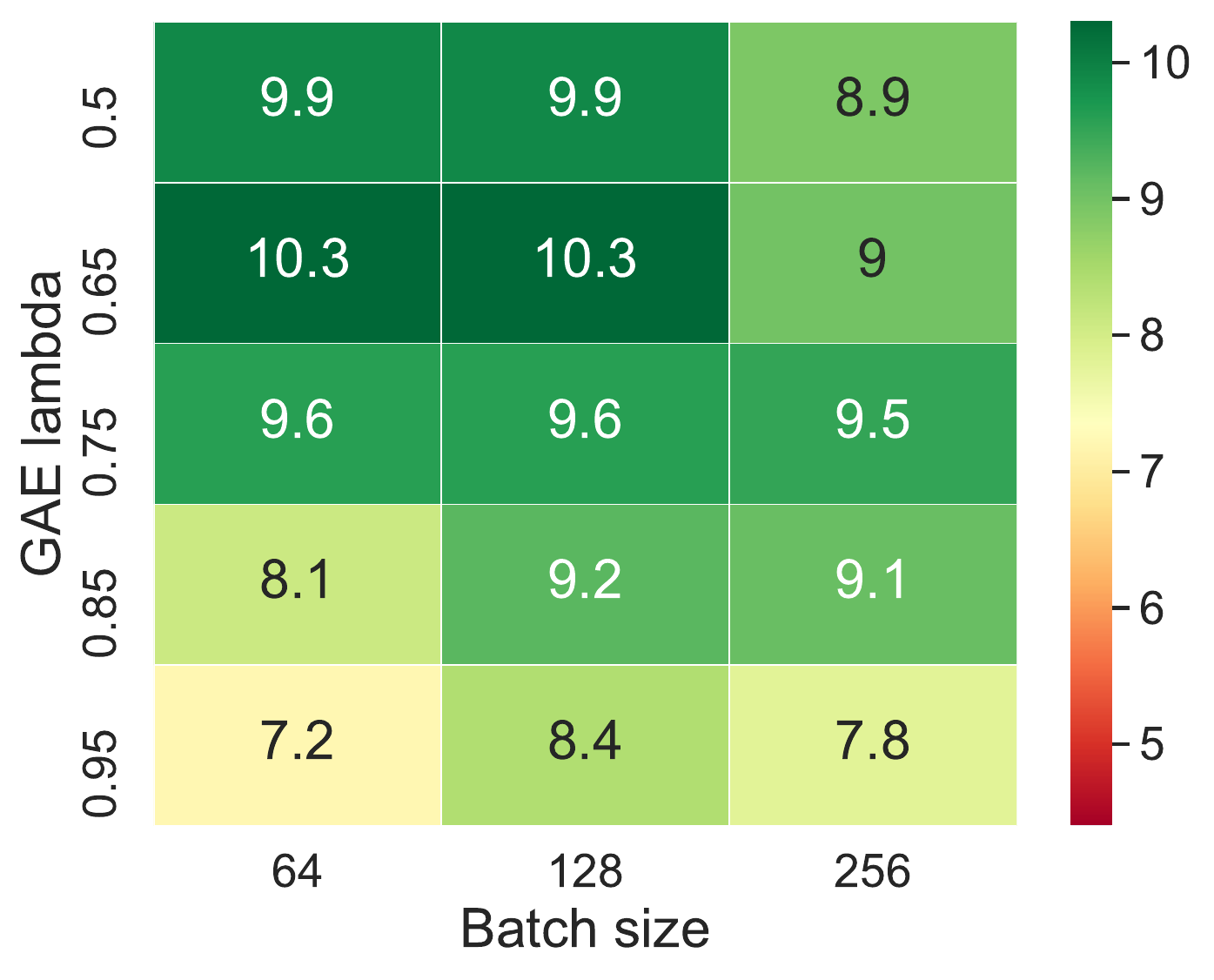}
        \label{sfig:hp-heatmap-cnnbest14}
	}
	\subfloat
	{
        \includegraphics[width=0.38\linewidth]{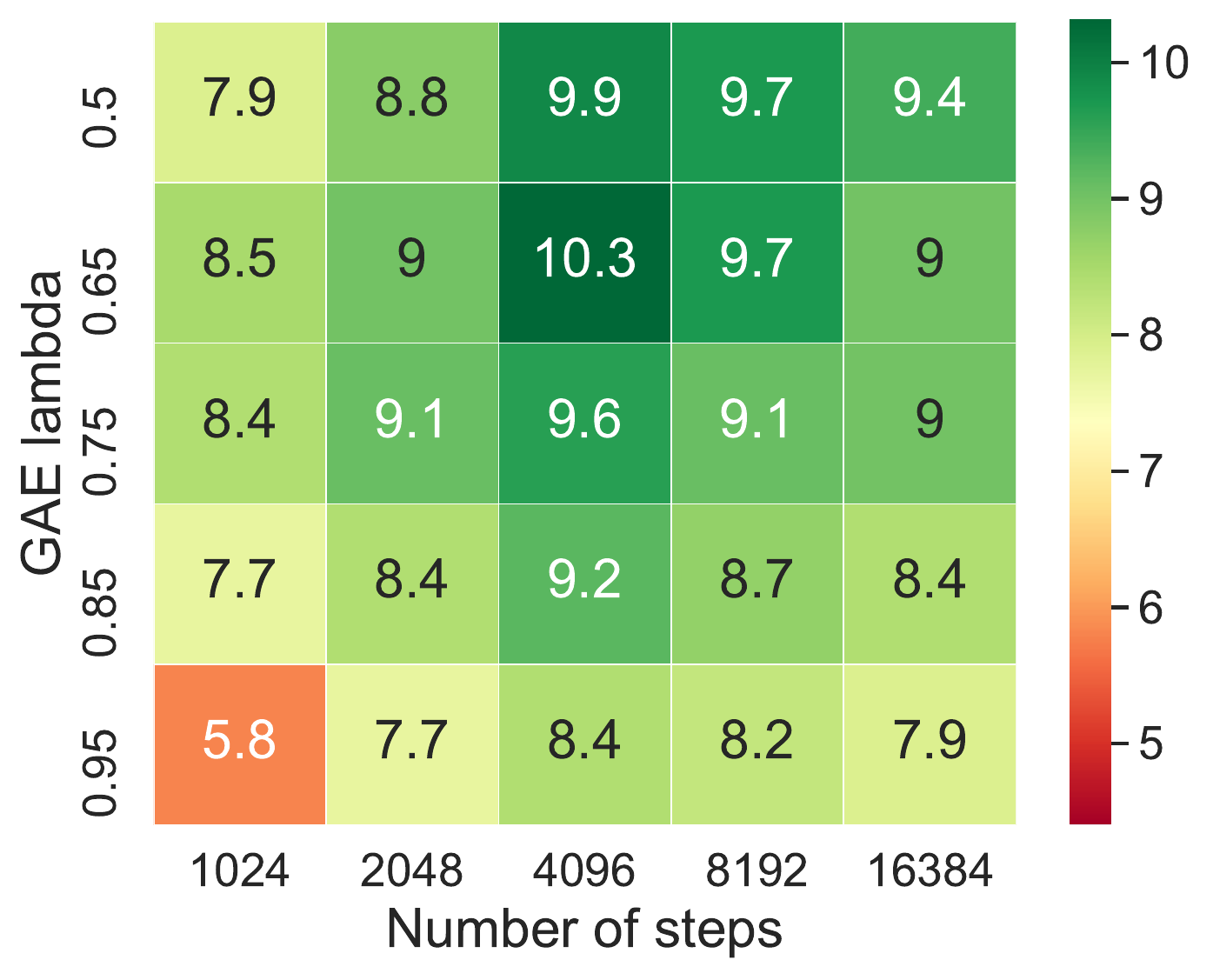}
        \label{sfig:hp-heatmap-cnnbest15}
	}
\hfil
	\subfloat
	{
        \includegraphics[width=0.38\linewidth]{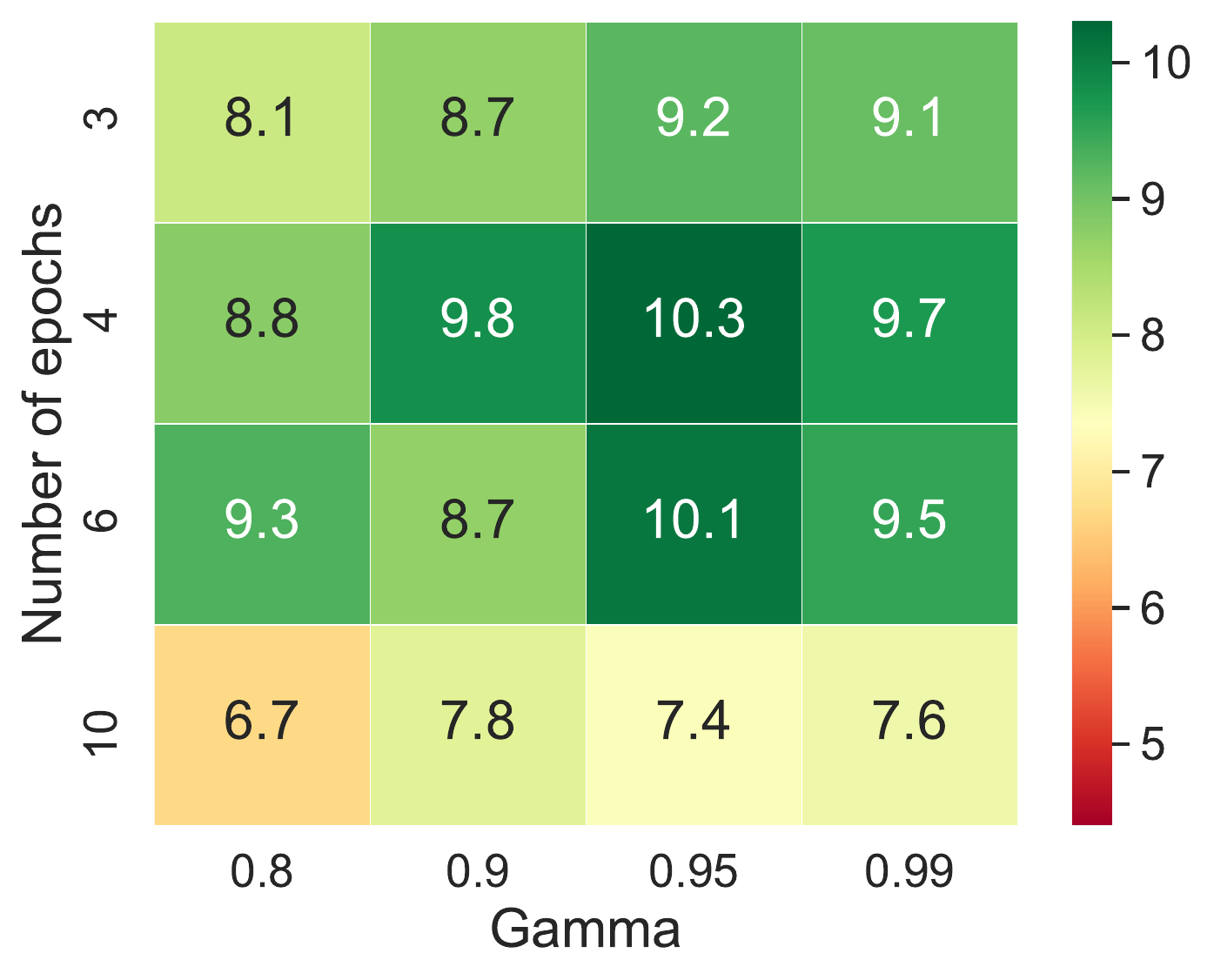}
        \label{sfig:hp-heatmap-cnnbest23}
	}
	\subfloat
	{
        \includegraphics[width=0.38\linewidth]{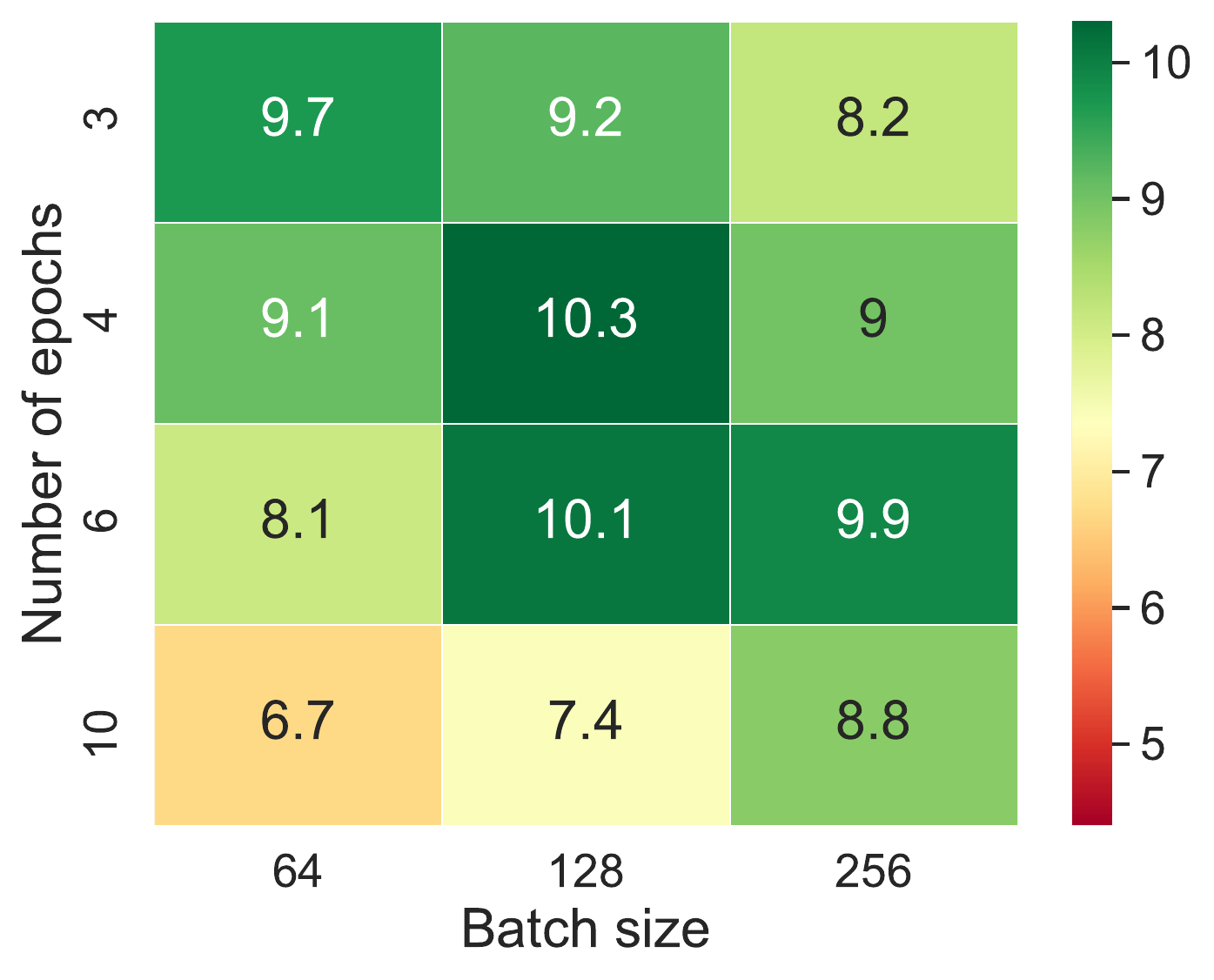}
        \label{sfig:hp-heatmap-cnnbest24}
	}
\hfil
	\subfloat
	{
        \includegraphics[width=0.38\linewidth]{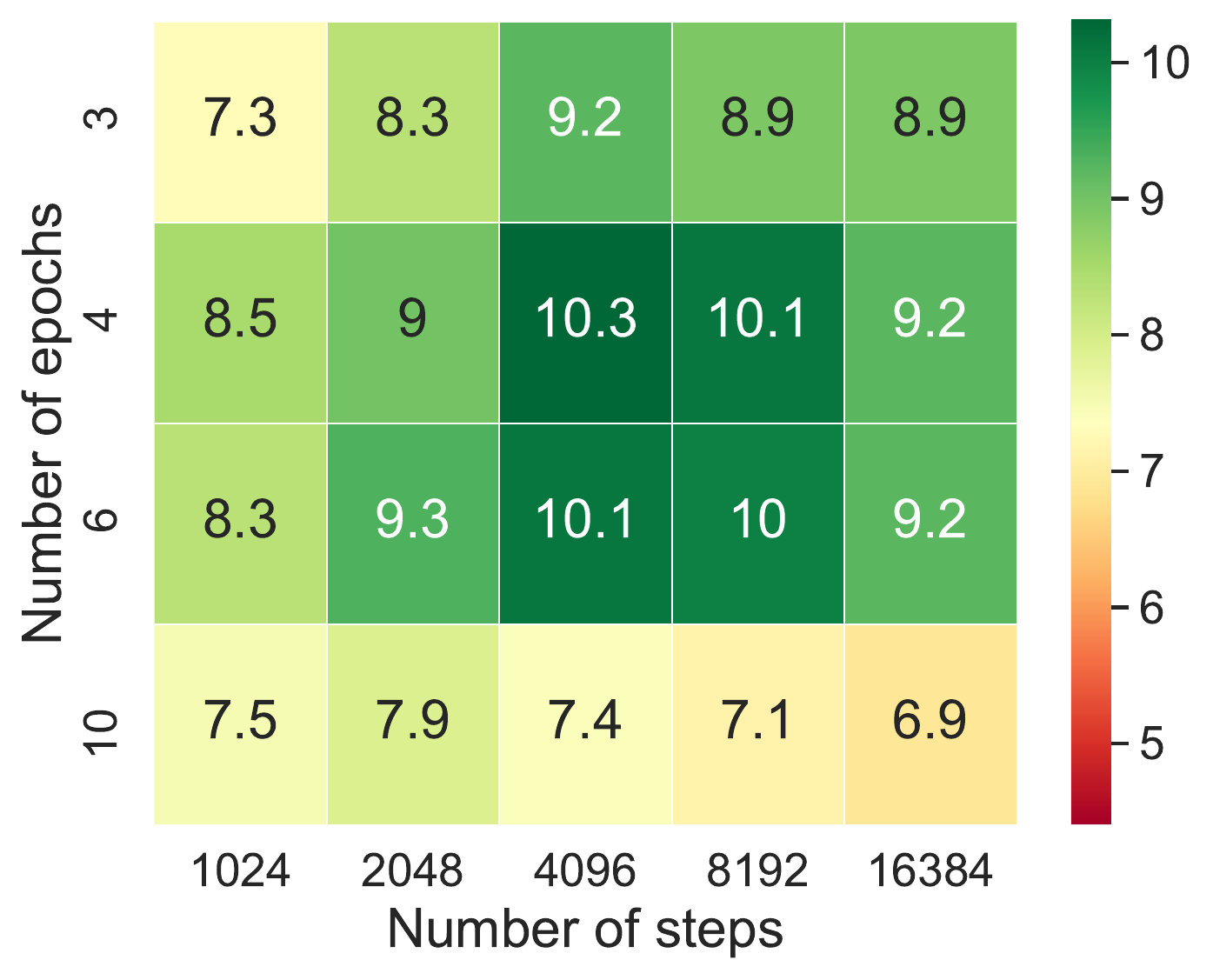}
        \label{sfig:hp-heatmap-cnnbest25}
	}
	\subfloat
	{
        \includegraphics[width=0.38\linewidth]{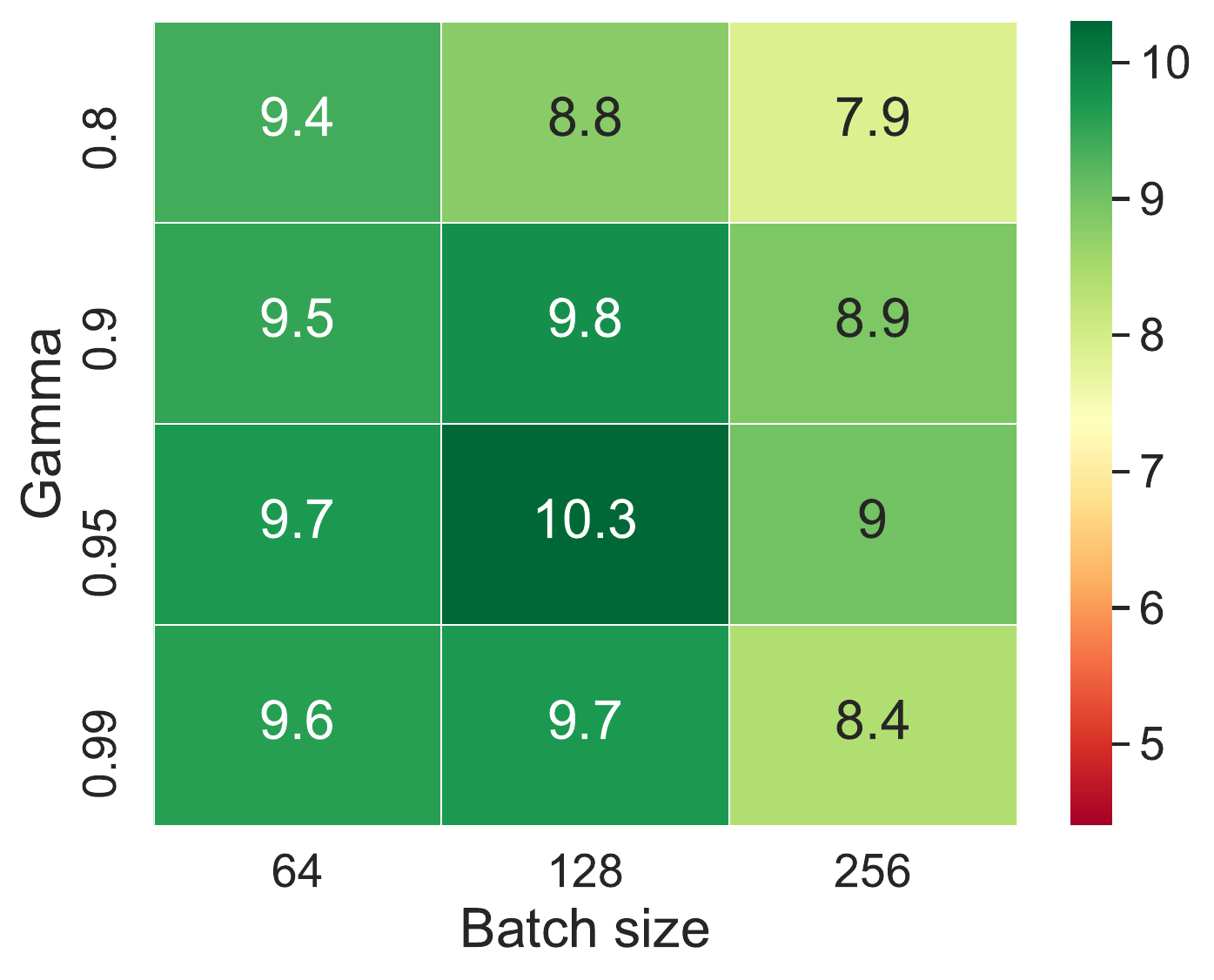}
        \label{sfig:hp-heatmap-cnnbest34}
	}
\hfil
	\caption{\it 
	\centering
    Crafter scores starting from the best hyper-parameters we found: GAE lambda 0.65, number of epochs 4, gamma 0.95, batch size 128 and number of steps 4096, and then ablating individual hyper-parameter pairs.
    }
	\label{fig:hp-heatmap-cnnbest}
\end{figure}

In \cref{fig:hp-heatmap-cnndef,fig:hp-heatmap-cnnbest} we show the pairwise heatmaps of the hyper-parameters we tuned for the PPO-CNN.
Important to note here is that once we tuned the hyper-parameters for PPO-CNN, we fixed them and used the same values for all other models.
Therefore, other models might have a slight disadvantage as we did not specifically tune hyper-parameters for each model.

In \cref{fig:hp-heatmap-cnndef} we start from the default hyper-parameters from \cite{stable-baselines3} that were tuned for Atari: GAE lambda 0.95, number of epochs 10, gamma 0.99, batch size 64 and number of steps 1024.
From here, we sweep over individual pairs of hyper-parameters, e.g. GAE lambda and the number of epochs.
Looking at the heatmaps, it is clear how each hyper-parameters contributes to the improvement (which becomes evident by looking at their combination in \cref{fig:hp-heatmap-cnnbest}). However, some hyper-parameters contribute much more than others, e.g. GAE lambda and the number of epochs the agent is trained on the collected rollouts.
Given these plots, we settled on the following values for hyper-parameters: number of steps 4096, number of epochs 4, batch size 128, GAE lambda 0.65, and gamma 0.95.
GAE-lambda can be interpreted as an extra discount factor applied after performing reward shaping transformation on the MDP \cite{schulman2015high}, and we found low values to perform better.
This also makes sense given that we found lower discount factors also to give better performance (0.95 compared to the most commonly used 0.99).
On the other hand, we extended the horizon (number of rollout steps collected for the update) from default 1024 to 4096) and similarly increased batch size from 64 to 128, stabilizing training agents in Crafter.
In general, when there are frequent rewards within an episode, the number of steps can be smaller. However, in Crafter, the rewards are sparse, which is also reflected in empirically better performance when using longer rollouts.

In \cref{fig:hp-heatmap-cnndef} we start from the best hyper-parameters we found: number of steps 4096, number of epochs 4, batch size 128, GAE lambda 0.65 and gamma 0.95.
We then ablate pairs of hyper-parameters by changing them over their respective ranges.
The agents are fairly robust to hyper-parameters changes. From the heatmaps, we can see that the performance drops going away from the optimal values, though not by a significant value.

\end{document}